\documentclass[a4paper]{article}
\usepackage{graphicx}
\usepackage[utf8]{inputenc}
\usepackage[russian]{babel}
\usepackage{amsfonts}
\usepackage[parfill]{parskip}
\usepackage{latexsym}
\usepackage{amsmath}
\usepackage{amssymb}
\usepackage{tikz}	
\usepackage{mathtools}

\mathtoolsset{showonlyrefs}
\usepackage{setspace}
\usepackage[left=3cm,right=1cm,
    top=2cm,bottom=2cm,bindingoffset=0cm]{geometry}
\usepackage{listings}
\usepackage{balance}
\usepackage{hyperref}
\usepackage{url}
\usepackage[square,numbers]{natbib}
\usepackage{ulem}
\usepackage{graphicx}
\usepackage{multirow}
\usepackage{array}
\usepackage{caption}
\usepackage[font=large]{caption}
\DeclareCaptionLabelFormat{dash}{Рисунок #2}
\DeclareCaptionFormat{listing}{#1 --- #3}
\usepackage{algpseudocode}
\usepackage{enumitem}
\usepackage{url}
\usepackage{pdfpages}

\usepackage[explicit]{titlesec}
\usepackage[nottoc]{tocbibind}
\usepackage{blindtext}

\begin{document}

\sloppy


\large
\doublespacing

\setlength{\parindent}{1.25cm}
\setlength{\parskip}{0cm}


\includepdf[pages={1}]{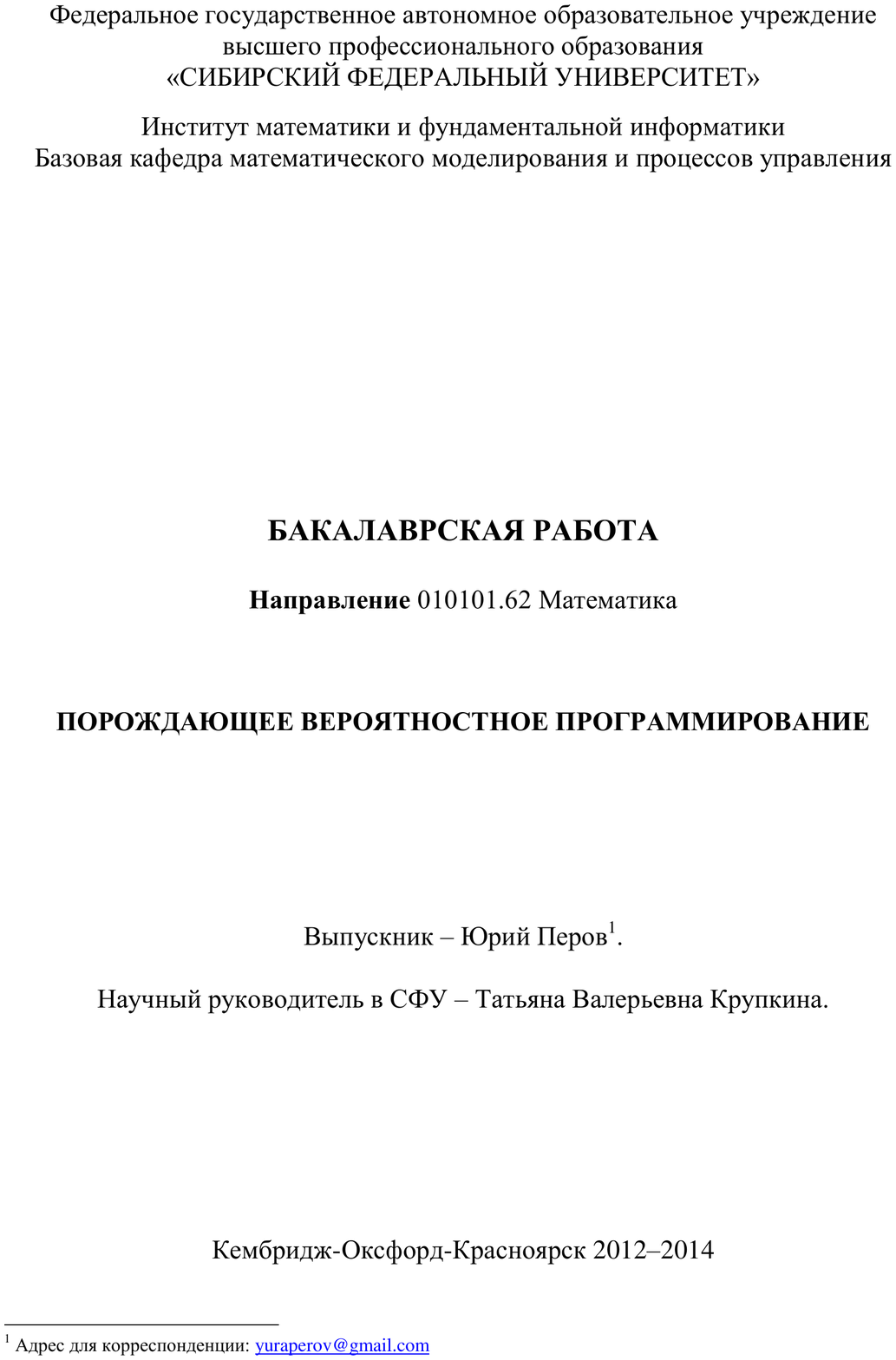}

\thispagestyle{empty}

\begingroup
\renewcommand{\section}[2]{}
\section*{РЕФЕРАТ}
\begin{center}
{\centering \Large \bf РЕФЕРАТ}
\end{center}
\vspace{5mm}
\endgroup

Выпускная квалификационная бакалаврская работа по теме <<Порождающее вероятностное программирование>> содержит 48 страниц текста, 50 использованных источников, 12 рисунков.

Ключевые слова: ВЕРОЯТНОСТНОЕ ПРОГРАММИРОВАНИЕ, МАШИННОЕ ОБУЧЕНИЕ, ИСКУССТВЕННЫЙ ИНТЕЛЛЕКТ, ВЕРОЯТНОСТНЫЕ МОДЕЛИ, АВТОМАТИЧЕСКОЕ МОДЕЛИРОВАНИЕ.

Работа посвящена новому направлению в области машинного обучения и компьютерных наук --- вероятностному программированию. В работе дается краткое реферативное введение в языки вероятностного программирования Church/Venture/Anglican, а также описываются результаты первых экспериментов по автоматической генерации вероятностных программ.

\clearpage

\setcounter{page}{2}

\begingroup
\titleformat{\section}[display]
{\normalfont\huge\bfseries}
{\chaptertitlename\ \thechapter}
{20pt}
{\Large \bf #1}
\titleformat{name=\section,numberless}[display]
{\normalfont\huge\filcenter\bfseries}
{}
{20pt}
{\Large\MakeUppercase{#1}}
\renewcommand\contentsname{СОДЕРЖАНИЕ}
\tableofcontents
\endgroup

\clearpage

\hfill \textit{Моей любимой Маше и всем}

\hfill \textit{другим моим любимым родным}

\hfill

\begingroup
\renewcommand{\section}[2]{}
\section*{ВВЕДЕНИЕ}
\addcontentsline{toc}{section}{Введение}
\begin{center}
{\centering \Large \bf ВВЕДЕНИЕ}
\end{center}
\vspace{5mm}
\endgroup

Данная бакалаврская работа посвящена вероятностному программированию \cite{goodman2013principles}, новому направлению в областях машинного обучения, искусственного интеллекта и компьютерных наук, а именно реферативному краткому введению в вероятностное программирование, описанию языков вероятностного программирования Church \cite{goodman2008church}, Venture \cite{venture} и Anglican \cite{wood2014new}, описанию подхода порождающего вероятностного программирования для решения задач распознавания образов \cite{mansinghka2013approximate}, а также представлению полученных предварительных результатов по автоматизации вывода вероятностных моделей для вероятностного программирования \cite{learning_probabilistic_programs}.

\subsection*{Описание структуры работы}

Первая часть данной работы начинается с общих сведений о вероятностном программировании, а затем в ней кратко описываются результаты, полученные коллегами автора и им самим в работе над научными проектами на протяжении двух лет в Массачусетском технологическом инситуте под руководством профессора Джошуа Тененбаума и доктора Викаша Мансингхи (Кембридж, штат Массачусетс, США) и Оксфордском университете под руководством профессора Френка Вуда (Великобритания, Оксфорд). Эта часть работы, являясь по сути переводом, представляет собой реферативную выдержку о вероятностном программировании и его приложениях на русском языке. Насколько автору известно, литературы о вероятностном программировании на русском языке практически нет, поэтому он надеется, что данная чисто реферативная часть работы принесет существенную пользу русскоговорящему научному сообществу, а особенно заинтересованным студентам и школьникам, которые впервые захотят познакомиться с развивающимся направлением вероятностного программирования.

Во второй части данной работы автор описывает новые результаты в области вероятностного программирования, связанные с автоматизированным или полуавтоматизированным выводом вероятностных моделей для вероятностного программирования, полученные во время стажировки автора в Департаменте технических наук Оксфордского университета в научной лаборатории профессора Френка Вуда и под его руководством. Вторая часть завершается рассуждениями автора об обобщении автоматизированного изучения и вывода компьютером вероятностных моделей в виде вероятностных программ, то есть о возможностях порождающего вероятностного программирования.

\subsection*{Вероятностное программирование}

Вероятностное программирование можно определить как компактный, композиционный способ представления порождающих вероятностных моделей и проведения статистического вывода в них с учетом данных с помощью обобщенных алгоритмов.

Вероятностная модель является важным понятием машинного обучения, одного из основных направлений искусственного интеллекта на сегодняшний день. В общем случае в рамках теории машинного обучения перед компьютером ставится задача произвести какое-то действие $y_i$ на основе входных данных $x_i$, априорных знаний и возможности взаимодействовать со средой. Без ограничений общности целевым действием для компьютера можно считать производство ответа в виде выходных данных, представленных в виде информации. Например, в робототехнике эта информация может являться инструкциями для моторов и механических устройств для выполнения тех или иных физических действий роботом.

При использовании различных подходов в рамках машинного обучения используются модели, которые являются формальным <<описанием компонентов и функций, отображающих существенные свойства моделируемого объема или процесса>> \cite{lopatnikov2003}. В рамках машинного обучения используются {\it вероятностные} модели, так как свойства, элементы и связи между ними являются не фиксированными, а стохастическими.

Классическим примером одного из подходов в машинном обучении является способ <<обучение с учителем>>, когда существует N прецедентов, то есть пар входных и выходных данных обучающей выборки $X = \{x_i\}, Y = \{y_i\}, i = \overline{1, N}$, и необходимо найти алгоритм описывающий зависимость между $x_i$ и $y_i$, то есть алгоритм $F$, который позволяет на основе каждого элемента входных данных $x_i$ получать абсолютно или достаточно точный элемент выходных данных $y_i$, то есть $F(x_i) \rightarrow y_i$. С помощью данного алгоритма затем для M известных элементов входных данных $\{x_i\}, j=\overline{N+1, N+M}$ находятся значения M неизвестных элементов выходных данных $\{y_i\}, j=\overline{N+1, N+M}$. В рамках машинного обучения эта проблема, в том числе, решается с помощью методов регрессионного анализа.

В только что приведенной как пример модели неизвестными (скрытыми) параметрами $T = \{t_j\}, j=\overline{1, K}$ будет информация о характеристиках модели $F$, которые подлежат выводу при данной обучающей выборке в виде пар $\left(x_i, y_i\right)$.

Приведем простой пример: в линейной регрессии (при $L$ независимых переменных) значениями независимых переменных будут $x_{i,j}$, значениями зависимой переменной будут $y_i$, параметрами модели будут $t_1, t_2, \ldots, t_{L+1}$, а алгоритмом $F$ будет
\begin{equation}
y_i = F_{t_1, \ldots, t_{L+1}}(x_i) = t_1 + t_2 \cdot x_{i,1} + \ldots + t_{L+1} \cdot x_{i,L}.
\label{eqn:1}
\end{equation}

Для простоты без ограничений общности будем считать, что мы можем всегда получить $y_i$ детерминированно, зная $T$ и $x_i$.

В порождающих вероятностных моделях задается совместное распределение вероятностей $P(T, X)$, обычно сначала путем задания априорного распределения $P(T)$, а затем задания условного распределения $P(X\ |\ T)$. Это и называется моделью.

При заданной модели и известных $X$ задачей будет являться поиск апостериорного распределения на $T$, таким образом $P(T\ |\ X)$. Одним из способов поиска данного апостериорного распределения является применение теоремы Байеса:

$$P(T\ |\ X) = \frac{P(T) P(X\ |\ T)}{P(X)},$$
где $P(X)$ теоретически можно найти как $\int { P(T) P(X\ |\ T)\ dT}$, но при решении практических задач это часто невозможно, так как перебор всего пространства T не поддается аналитическому решению или решению с помощью численных методов за разумное время. Поэтому чаще всего при решении задач в рамках Байесовского подхода стоит задача поиска ненормированного значения $P(T\ |\ X)$: $$P(T\ |\ X) \propto P(T) P(X\ |\ T),$$
где символ $\propto$, часто встречающийся в зарубежной литературе, но очень редко встречающийся у нас, означает <<пропорционально>>. Нормировочную константу затем можно найти приближенно, но иногда ее значение даже не вычисляют, так как бывает достаточно найти и работать дальше с наиболее вероятными элементами $\hat{T}$ из апостериорного распределения $P(T\ |\ X)$.

Обычно, особенно при решении практических задач с большим объемом данных и в рамках сложных моделей, апостериорное распределение $P(T\ |\ X)$ находят не точно, а с помощью приближенных методов, в том числе с помощью методов Монте-Карло \cite{russian_monte_carlo_methods}, которые позволяют сгенерировать выборку из интересующего нас распределения.

Как отмечалось в самом начале данного подраздела, вероятностное программирование позволяет:
\begin{enumerate}[leftmargin=1.75cm]
\item Композиционно и компактно записывать порождающую вероятностную модель с помощью задания априорных вероятностей $P(T)$ и $P(X\ |\ T)$ в виде алгоритма (вероятностной программы) с использованием стохастических функций (например, стохастическая функция $\mathrm{Normal}(\mu, \sigma)$) и вспомогательных детерминированных функций (например, $\mathrm{+}$ или $\mathrm{*}$).
\item Снабжать модель данными $\hat{X}$, таким образом теоретически определяя условное распределение $P(T\ |\ \hat{X})$.
\item Производить статистический вывод для генерации выборки из условного распределения  с помощью обобщенных алгоритмов статистического вывода, подходящих для всех или большого множества моделей. 
\end{enumerate}

Для ознакомления с машинным обучением и искусственным интеллектом автор рекомендует следующие источники: \cite{russell_artificial,bishop2006pattern,murphy2012machine}. Информацию о вероятностных моделях и Байесовских методах на русском языке можно найти в \cite{bayesian_methods_of_machine_learning}.

\clearpage

\section{Краткое введение в языки вероятностного программирования Church, Venture и Anglican}

Существует более 15 языков вероятностного программирования, перечень с кратким описанием каждого из них можно найти на \cite{pp_site}. В данной работе реферативно будут рассмотрены три языка вероятностного программирования: Church \cite{goodman2008church}, Venture \cite{venture} и Anglican \cite{wood2014new}. Языки Venture и Anglican являются продолжениями языка Church. Church в свою очередь основан на языке <<обычного>> программирования Lisp и Scheme. Заинтересованному читателю крайне рекомендуется ознакомиться с книгой \cite{Russian_SICP}, являющейся одним из лучших способов начать знакомство с языком <<обычного>> программирования Scheme.

\subsection{Первое знакомство с Church, Venture, Anglican}

На текущий момент любой язык вероятностного программирования напрямую связан с методами статистического вывода, которые используются в нем, поэтому часто язык ассоциируется с платформой вероятностного программирования, то есть с его технической реализацией в виде компьютерной программы.

Рассмотрим задание простой вероятностной модели Байесовской линейной регрессии \cite{bishop2006pattern} на языке вероятностного программирования Venture/Anglican \cite{roy_introducing_linear_regression} в виде вероятностной программы:

\lstset{
  numbers=left
}
\begin{lstlisting}{}{program:abc}
[ASSUME t1 (normal 0 1)]
[ASSUME t2 (normal 0 1)]
[ASSUME noise 0.01]
[ASSUME noisy_x (lambda (time) (normal (+ t1 (* t2 time)) noise))]
[OBSERVE (noisy_x 1.0) 10.3]
[OBSERVE (noisy_x 2.0) 11.1]
[OBSERVE (noisy_x 3.0) 11.9]
[PREDICT t1]
[PREDICT t2]
[PREDICT (noisy_x 4.0)]
\end{lstlisting}

Скрытые искомые параметры --- значения коэффициентов $t_1$ и $t_2$ линейной функции $x\left(time\right) = t_1 + t_2 \cdot time$. У нас есть априорные предположения о данных коэффициентах, а именно мы предполагаем, что они распределены по закону нормального распределения $\mathrm{Normal}(0, 1)$ со средним $0$ и стандартным отклонением $1$. Таким образом, мы определили в первых двух строках вероятностной программы вероятность $P(T)$, описанную в предыдущем раздел. Инструкцию \texttt{[ASSUME name expression]} можно рассматривать как определение случайной величины с именем \texttt{name}, принимающей значение вычисляемого выражение (программного кода) \texttt{expression}, которое содержит в себе неопределенность.

Вероятностные языки программирования (здесь и далее будут иметься в виду конкретно Church, Venture, Anglican, если не указано иное), как и Lisp/Scheme, являются функциональными языками программирования, и используют польскую нотацию\footnote{Venture имеет отдельный дополнительный вид синтаксиса VentureScript, использующий инфиксную нотацию и не требующий обрамления вызова функций скобками, то есть схожий по своей сути с привычными большинству людей языками программирования C, C++, Python и т.д.} при записи выражений для вычисления. Это означает, что в выражении вызова функции сначала располагается оператор, а уже только потом аргументы: \texttt{(+ 1 2)}, и вызов функции обрамляется круглыми скобками. На других языках программирования, таких как C++ или Python, это будет эквивалентно коду \texttt{1 + 2}.

В вероятностных языках программирования выражение вызова функции принято разделять на три разных вида:
\begin{enumerate}[leftmargin=1.75cm]
\item Вызов детерминированных процедур \texttt{(primitive-procedure arg1 \ldots argN)}, которые при одних и тех же аргументах всегда возвращают одно и то же значение. К таким процедурам, например, относятся арифметические операции.
\item Вызов вероятностных (стохастических) процедур \texttt{(stochastic-procedure arg1 \ldots argN)}, которые при каждом вызове генерируют случайным образом элемент из соответствующего распределения. Такой вызов определяет новую {\it случайную величину}. Например, вызов вероятностной процедуры \texttt{(normal 1 10)} определяет случайную величину, распределенную по закону нормального распределения $\mathrm{Normal}(1, \sqrt{10})$, и результатом выполнения каждый раз будет какое-то вещественное число.
\item Вызов составных процедур \texttt{(compound-procedure arg1 \ldots argN)}, где \texttt{compound-procedure} --- введенная пользователем процедура с помощью специального выражения \texttt{lambda}: \texttt{(lambda (arg1 \ldots argN) body)}, где \texttt{body} --- тело процедуры, состоящее из выражений. В общем случае составная процедура является стохастической (недетерминированной) составной процедурой, так как ее тело может содержать вызовы вероятностных процедур.
\end{enumerate}

После этого мы хотим задать условную вероятность $P(X\ |\ T)$ наблюдаемых переменных $x_1, x_2, x_3$ при заданных значениях скрытых переменных $t_1, t_2$ и параметра $time$.

Перед вводом непосредственно самих наблюдений с помощью выражения \texttt{[OBSERVE \ldots]} мы определяем общий закон для наблюдаемых переменных $\{x_i\}$ в рамках нашей модели, а именно мы предполагаем, что данные наблюдаемые случайные величины при заданных $t_1, t_2$ и заданном уровне шума $noise$ распределены по закону нормального распределения $\mathrm{Normal}(t_1 + t_2 \cdot time, \sqrt{noise})$ со средним $t_1 + t_2 \cdot time$ и стандартным отклонением $noise$. Данная условная вероятность определена на строках 3 и 4 данной вероятностной программы. \texttt{noisy\_x} определена как функция, принимающая параметр $time$ и возвращающая случайное значение, определенное с помощью вычисления выражение \texttt{(normal (+ t1 (* t2 time)) noise)} и обусловленное значениями случайных величин $t_1$ и $t_2$ и переменной $noise$. Отметим, что выражение \texttt{(normal (+ t1 (* t2 time)) noise)} содержит в себе неопределенность, поэтому каждый раз при его вычислении мы будем получать в общем случае разное значение.

На строках 5---7 мы непосредственно вводим известные значения $\hat{x_1} = 10.3$, $\hat{x_2} = 11.1$, $\hat{x_3} = 11.9$. Инструкция вида \texttt{[OBSERVE expression value]} фиксирует наблюдение о том, что случайная величина, принимающая значение согласно выполнению выражения \texttt{expression}, приняла значение \texttt{value}.

Повторим на данном этапе всё, что мы сделали. На строках 1---4 с помощью инструкций вида \texttt{[ASSUME \ldots]} мы задали непосредственно саму вероятностную модель: $P(T)$ и $P(X\ |\ T)$. На строках 5---7 мы непосредственно задали известные нам значения наблюдаемых случайных величин $X$ с помощью инструкций вида \texttt{[OBSERVE \ldots]}.

На строках 8---9 мы запрашиваем у системы вероятностного программирования апостериорное распределение $P(T\ |\ X)$ скрытых случайных величин $t_1$ и $t_2$. Как уже было сказано, при большом объеме данных и достаточно сложных моделях получить точное аналитическое представление невозможно, поэтому инструкции вида \texttt{[PREDICT \ldots]} генерируют выборку значений случайных величин из апостериорного распределения $P(T\ |\ X)$ или его приближения. Инструкция вида \texttt{[PREDICT expression]} в общем случае генерирует один элемент выборки из значений случайной величины, принимающей значение согласно выполнению выражения \texttt{expression}. Если перед инструкциями вида \texttt{[PREDICT \ldots]} расположены инструкции вида \texttt{[OBSERVE ...]}, то выборка будет из апостериорного распределения\footnote{Говоря точнее, конечно, из приближения апостериорного распределения.}, обусловленного перечисленными ранее введенными наблюдениями.

Отметим, что в завершении мы можем также предсказать значение функции $x(time)$ в другой точке, например, при $time = 4.0$. Под предсказанием в данном случае понимается генерация выборки из апостериорного распределения новой случайной величины при значениях скрытых случайных величин $t_1, t_2$ и параметре $time = 4.0$.

Для генерации выборки из апостериорного распределения $P(T\ |\ X)$ в языке программирования Church в качестве основного используется алгоритм Метрополиса-Гастингса, который относится к методам Монте-Карло по схеме Марковских цепей. В следующем подразделе будет произведено подробное описание применения данного алгоритма для {\it обобщенного} статистического вывода в вероятностных языках. Под <<обобщенным>> выводом в данном случае понимается то, что алгоритм может быть применен к любым вероятностным программам, написанным на данном вероятностном языке программирования.

\subsection{Статистический вывод в вероятностных языках программирования с помощью алгоритма Метрополиса-Гастингса}

Описание алгоритма Метрополиса-Гастингса в применении к <<семейству>> вероятностных языков Church впервые опубликовано в \cite{goodman2008church} и более подробно описано в \cite{wingate2011lightweight}.

Получить выборку из $N$ элементов из априорного распределения скрытых параметров $P(T)$, наблюдаемых величин $P(X\ |\ T)$ или их совместного распределение $P(T, X)$ какой-либо порождающей вероятностной модели, записанной в виде вероятностной программы, не составляет труда. Для этого достаточно выполнить вероятностную программу $N$ раз. Отметим очевидный факт, что в данном случае вероятностная программа будет содержать лишь инструкции вида \texttt{[ASSUME ...]} (задание вероятностной модели) и \texttt{[PREDICT ...]} (перечисление случайных величин, выборку которых мы генерируем).

\subsubsection{Метод <<выборки с отклонением>>}

При заданных наблюдениях с помощью инструкций вида \texttt{[OBSERVE expression value]} наиболее простым способом получения апостериорного распределения является метод <<выборки с отклонением>> \cite{nikolenko_probabilistic_learning}. Для понимания рассмотрим следующую вероятностную программу:

\lstset{
  numbers=left
}
\begin{lstlisting}{}{program:abc}
[ASSUME a (uniform-discrete 1 6)]
[ASSUME b (uniform-discrete 1 6)]
[OBSERVE (+ a b) 5]
[PREDICT (* a b)]
\end{lstlisting}

Отметим, что стохастическая процедура \texttt{(uniform-continuous a b)} возвращает значение случайной величины, распределенной по равномерному дискретному закону распределения с носителем $\{a, \ldots, b\}$.

Текстом задачу, записанную выше с помощью вероятностной программы, можно сформулировать следующим образом: подбрасываются два шестигранных игральных кубика, в сумме выпало 5; каково распределение произведения очков на этих двух кубиках?

Метод <<выборки с отклонением>> заключается в том, чтобы генерировать значения очков на первом и втором кубиках из их априорного распределения (таким образом, два независимых дискретных равномерных распределения) и проверять, равна ли их сумма 5. Если нет, данная попытка отвергается и не учитывается. Если да, то произведение очков на кубиках добавляется во множество элементов выборки. Данная выборка и будет аппроксимацией апостериорного распределения значений произведения очков на двух кубиках, если известно, что сумма очков равна 5.

Данный метод является неэффективным, и при решении более сложных задач просто вычислительно неосуществимым. Также отметим, что он хорошо подходит только для дискретных значений случайных и промежуточных переменных.

\subsubsection{Пространство историй выполнений вероятностных программ}

При выполнении вероятностной программы каждая случайная величина принимает определенное значение. Например, следующая вероятностная программа содержит три случайных величины, каждая из которых распределена по закону нормального распределения:

\lstset{
  numbers=left
}
\begin{lstlisting}{}{program:abc}
[ASSUME a (normal 0 1)]
[ASSUME b (normal 0 1)]
[ASSUME c (normal (+ a b) 1)]
[PREDICT c]
\end{lstlisting}

При выполнении данной вероятностной программы каждая из трех случайных величин примет свое случайное значение. Назовем {\it историей выполнения } вероятностной программы отображение, сопоставляющее каждой случайной величине ее значение. Мощность {\it истории выполнения} совпадает с количеством случайных величин, которые были созданы при ее выполнении.

Для каждой программы можно определить соответствующее вероятностное пространство, элементарными событиями которого будут являться все {\it истории ее выполнения}. Вероятностью каждого элементарного события будет являться произведение вероятностей принятие того или иного значения каждой случайные величины при данной истории выполнения.

Очевидно, что история выполнения однозначным образом определяет выполнение вероятностной программы и принимаемые значения, как случайных, так и зависящих от них детерминированных выражений.

Для примера рассмотрим еще более простую вероятностную программу.

\lstset{
  numbers=left
}
\begin{lstlisting}{}{program:abc}
[ASSUME a (bernoulli 0.7)]
[ASSUME b (bernoulli 0.7)]
[PREDICT a]
[PREDICT b]
\end{lstlisting}

В ней обе случайных величины имеют свое название, что облегчает запись историй выполнений (иначе необходимо вводить адресную схему для наименования случайных величин, чтобы однозначно идентифицировать их).

Данная вероятностная программа имеет четыре различных возможных историй выполнений, а именно $\{a: 0, b: 0\}$, $\{a: 0, b: 1\}$, $\{a: 1, b: 0\}$ и $\{a: 1, b: 1\}$ с вероятностями $0.09$, $0.21$, $0.21$ и $0.49$ соответственно.

Отметим также, что количество <<активных>> случайных величин при выполнении одной и той же вероятностной программы может меняться, как, например, в следующей программе:

\lstset{
  numbers=none
}
\begin{lstlisting}{}{program:abc}
[ASSUME geometric
  (lambda (p) (if (= (bernoulli p) 1) 0 (+ (geometric p) 1)))]
[PREDICT (geometric 0.5)]
\end{lstlisting}

Данная программа генерирует элемент (т.е. одноэлементную выборку) из геометрического распределения с параметром 0.5 с помощью составной процедуры \texttt{geometric}, параметризованной параметром $p$. При выполнении данной вероятностной программы в истории ее выполнения количество <<активных>> случайных величин может быть любым.

\subsubsection{Апостериорное распределение историй выполнений программ}

При добавлении наблюдений, то есть фиксации значения определенных случайных величин, можно считать, что рассматривается подмножество множества элементарных событий, а именно те элементы, в историях выполнений которых наблюдаемые случайные величины принимают желаемое значение.

Можно описать другое вероятностное пространство, множеством элементарных событий которого будет являться только что описанное подмножество. Вероятностная мера в новом вероятностном пространстве может быть <<индуцирована>> вероятностной мерой из первоначального вероятностного пространства с учетом нормировочной константы.

Только что описанный переход полностью сочетается с теоремой Байеса, которая в нашем случае записывается следующим образом:

$$P(T\ |\ X = \hat{x}) = \frac{P(T)P(X = \hat{x}\ |\ T)}{P(X = \hat{x})},$$
где $X$ --- множество случайных величин, для которых мы знаем фиксированные наблюдаемые значения; $T$ --- множество случайных величин, ассоциируемых со скрытыми параметрами, апостериорное распределение которых мы заинтересованы получить (грубо говоря, это всё остальные случайные величины); $\hat{x}$ --- наблюдаемые значения случайных величин.

Отметим, что часто мы заинтересованы не в апостериорном распределении $P(T\ |\ X = \hat{x})$, а в апостериорном распределении лишь части скрытых параметров $P(T' \subset T\ |\ X = \hat{x})$ или даже функции $f$ от них $P(f(T)\ |\ X = \hat{x})$, в общем случае не являющейся биекцией. С другой стороны, так как мы не ставим задачу получить аналитическое представление данных апостериорных распределений, а лишь выборку из них, то это не играет большой роли в нашем случае при решении задач методами Монте-Карло: мы можем генерировать элементы выборки из $P(T\ |\ X = \hat{x})$, и затем использовать только значения нужных нам скрытых случайных величин и/или действовать функцией $f$ на них.

\subsubsection{Использование методов Монте-Карло по схеме Марковских цепей}

Математическо-статистический аппарат методов Монте-Карло по схеме Марковских цепей кратко и <<современно>> изложен в \cite{russian_monte_carlo_methods}.

Интуитивно опишем, что мы собираемся делать. У нас есть вероятностная программа $g$, определяющая множество скрытых $T$ и наблюдаемых $X$ случайных величин. Вероятностная программа своей записью задает априорное распределение $P(T), P(X\ |\ T)$, а значит и совместное распределение $P(T, X)$. Каждый раз путем выполнения данной программы мы получаем одну из возможных реализаций данной программы, которая биективно описывается соответствующей историей выполнения $h_i \in H$, где множество $H$ --- множество всех возможных историй выполнений данной вероятностной программы, которое можно рассматривать как множество элементарных событий. Для каждой истории выполнения определена ее вероятностная мера $$P(h_i) = P(T = \tilde{t}, X = \tilde{x}) = P(T = \tilde{t}) P(X = \tilde{x}\ |\ T = \tilde{t}).$$

У нас также есть <<экспериментальное>> значение каждой наблюдаемой случайной величины $\hat{x}_i$, то есть множество $\{\hat{x}_i\} = \hat{x}$. Существует подмножество историй выполнений $H' \subset H$, в котором наблюдаемые случайные величины принимают желаемое значение. Данное подмножество можно рассматривать как множество элементарных событий другого вероятностного пространства, вероятностную меру на котором можно <<индуцировать>> из предыдущего путем деления на нормирующую постоянную $P(X = \hat{x})$:
$$P(h') = \frac{P(T = \tilde{t}) P(X = \hat{x}\ |\ T = \tilde{t})}{P(X = \hat{x})}.$$

Будем обозначать ненормированную вероятностную меру
$$\tilde{P(h')} = P(T = \tilde{t}) P(X = \hat{x}\ |\ T = \tilde{t}).$$

Рассмотрим цепь Маркова, исходами которой являются истории выполнений $h' \in H'$. В качестве начального распределения можно выбрать априорное распределение $P(T) P(X\ |\ T)$, при котором значения наблюдаемых случайных величин не выбираются случайным образом согласно их распределению, а устанавливаются согласно их значениям. Например, в вероятностной программе
\lstset{
  numbers=left
}
\begin{lstlisting}{}{program:abc}
[ASSUME a (gamma 1 1)]
[ASSUME b (lambda () (normal a 1))]
[OBSERVE (b) 5.3]
\end{lstlisting}
первая случайная величина $a$, распределенная по закону Гамма-распределения $\mathrm{Gamma}(1, 1)$, будет являться скрытой, и будет сгенерирована согласно данному закону распределения. Наблюдаемая же случайная величина, распределенная по нормальному закону $\mathrm{Normal}(a, 1)$, не будет сгенерирована, а будет установлена в соответствии с ее наблюдаемым значением.

Мы хотим установить такие правила перехода по схеме Метрополиса-Гастингса из одного состояния (исхода) цепи Маркова в другое, чтобы стационарное распределение данной цепи Маркова совпадало с распределением $P(h')$. В таком случае для получения аппроксимации искомого апостериорного распределения в виде выборки нам будет достаточно имитировать данную цепь Маркова \cite{russian_monte_carlo_methods,chib1995understanding}.

В алгоритме Метрополиса-Гастингса вероятностная мера может быть известна с точностью до нормировочной константы, что и происходит в нашем случае. На каждом шаге алгоритма дано текущее состояние $h'_t$ и в соответствии с заданным заранее условным распределение предлагается новое состояние $h'^{*} \sim Q(\ \cdot \ |\ h'_t)$. Таким образом, $Q$ можно назвать распределением предлагаемых переходов. После этого подсчитывается коэффициент <<принятия>> нового состояния:

$$\alpha = \min \left( 1, \frac{\tilde{P}(h'^{*})\ Q(h'_t\ |\ h'^{*})}{\tilde{P}(h'_t)\ Q(h'^{*}\ |\ h'_t)} \right).$$

Состояние $h'^{*}$ принимается в качестве следующего состояния $h'_{t+1}$ с вероятностью $\alpha$, в противном случае $h'_{t+1} := h'_t$.

Новое состояние предлагается следующим образом:
\begin{enumerate}[leftmargin=1.75cm]
\item Случайным образом (равномерно) выбирается одна <<активная>> скрытая случайная величина $r \in \operatorname{dom} h'_t$ для <<вариации>>.
\item Предлагается новое значение данной случайной величины $r^{*} \sim \kappa(\ \cdot\ |\ r = \hat{r}, h'_t)$. Условная вероятность $\kappa$ в данном случае будет локальным для $r$ распределением предлагаемых переходов.
\item Если случайная величина $r$ влияет на поток выполнения вероятностной программы, и в результате ее нового значения должны быть исполнены другие ветви выполнения вероятностной программы, они исполняются и в общем случае происходит генерация новых случайных величин, которые прежде были неактивны.
\end{enumerate}

Для примера рассмотрим следующую вероятностную программу:
\lstset{
  numbers=left
}
\begin{lstlisting}{}{program:abc}
[ASSUME a (bernoulli 0.3)]
[ASSUME b (if (= a 1) (normal 0 1) (gamma 1 1))]
[PREDICT c]
\end{lstlisting}
В данной вероятностной программе три случайных величины, первая $\xi_1$ распределена по закону Бернулли, вторая $\xi_2$ по закону нормального распределения, третья $\xi_3$ по закону Гамма-распределения. $\xi_1$ можно также называть <<a>> (по имени переменной), хотя переменная <<b>> не является сама по себе случайной величиной, а зависит от значения $\xi_1$, определяющей поток выполнения вероятностной программы, и от значения либо $\xi_2$, либо $\xi_3$. Отметим также, что при каждом выполнении данной вероятностной программы будет существовать только две {\it активных} случайных величины.

Предположим, что текущим состоянием вероятностной программы в момент времени $t$ была история выполнения $h'_t = \{\xi_1 = 1, \xi_2 = 3.2\}$. Это означает, что случайная величина $\xi$ приняла значение $1$, и поэтому для задания переменной $b$ был сгенерирован элемент из нормального распределения, то есть была <<реализована>> случайная величина $\xi_2 \sim \mathrm{Normal}(0, 1)$. Очевидно, что
$$P(h'_t) = P(\xi_1 = 1) \cdot P(\xi_2 = 3.2\ |\ \xi_1 = 1) = 0.3 \cdot f_{\mathrm{Normal}(0, 1)}(3.2) \approx 0.0007152264603.$$

Предложим новое состояние $h'^{*}$:
\begin{enumerate}[leftmargin=1.75cm]
\item Выберем случайным образом одну из двух <<активных>> случайных величин, пусть это будет $\xi_1$.
\item Предложим случайным образом новое значение данной случайной величины согласно ее априорному распределению (то есть $\mathrm{Bernoulli}(0.3)$), пусть это будет $0$.
\item Так как случайная величина $\xi_1$ действительно влияет на поток выполнения вероятностной программы и при изменении ее значения в данном случае должна быть выполнена другая ветвь, выполним данную ветвь, генерируя значения <<активирующихся>> случайных величин (в нашем случае только $\xi_3$). Предположим, что $\xi_3$ стало равным $12.3$.
\end{enumerate}

Тогда $$Q(h'^{*}\ |\ h'_t) = \frac{1}{2} \cdot 0.7 \cdot f_{\mathrm{Gamma}(1, 1)}(12.3).$$

В общем случае
$$Q(h'^{*}\ |\ h'_t) = \frac{1}{|h'_t|} \cdot \kappa(r^{*}\ |\ r = \hat{r}, h'_t) \cdot R_{new},$$
где $|h'_t|$ --- количество активных случайных величин в текущей истории выполнения, $\kappa(r^{*}\ |\ r = \hat{r}, h'_t)$ --- вероятность нового значения варьируемой случайной величины $r$, $R_{new}$ --- совместная вероятность выбора своих значений <<активирующихся>> случайных величин.

При уже фиксированном $h'^{*}$ обратное $Q(h'_t\ |\ h'^{*})$ для коэффициента принятия алгоритма Метрополиса-Гастингса в общем случае может быть посчитано аналогичным образом:
$$Q(h'_t\ |\ h'^{*}) = \frac{1}{|h'^{*}|} \cdot \kappa(\hat{r}\ |\ r = r^{*}, h'^{*}) \cdot R_{old},$$
где $R_{old}$ --- совместная вероятность выбора своих значений случайных величин, активных в $h'_t$, но не являющихся активными в $h'^{*}$.

В только что рассмотренном примере
$$Q(h'_t\ |\ h'^{*}) = \frac{1}{2} \cdot 0.3 \cdot f_{\mathrm{Normal}(0, 1)}(3.2).$$

И тогда с учетом того, что $P(h'^{*}) = 0.7 \cdot f_{\mathrm{Gamma}(1, 1)}(12.3) \approx 0.000003186221124$, мы получаем, что 
$$\alpha = \min\left(1, \frac{0.7 \cdot f_{\mathrm{Gamma}(1, 1)}(12.3) \cdot \frac{1}{2} \cdot 0.3 \cdot f_{\mathrm{Normal}(0, 1)}(3.2)}{0.3 \cdot f_{\mathrm{Normal}(0, 1)}(3.2) \cdot \frac{1}{2} \cdot 0.7 \cdot f_{\mathrm{Gamma}(1, 1)}(12.3)}\right) = 1.$$

Поэтому в данном конкретном примере $h'_{t+1} = h'^{*}$ в любом случае, то есть с вероятностью один.

Выбор локального для $r$ распределения предлагаемых переходов может быть разным. В простейшем случае, если $r$ --- независимая случайная величина, данное распределение выбирается идентичным априорному распределению для $r$. Если же $r$ --- перестановочная случайная величина \cite{zubkov_erp}, а такие случайные величины поддерживаются рассматриваемыми языками вероятностного программирования, то ее распределение может быть выбрано с учетом уже накопленных значений.

Таким образом, мы описали алгоритм для предложения нового состояния $h'^{*}$ при текущем состоянии $h'_t$ и описали его в рамках метода Метрополиса-Гастингса. Чтобы получить выборку $\{h'_j\}_{j=1}^{N}$ из $N$ элементов желаемого апостериорного распределения, нам необходимо имитировать данную цепь Маркова согласно описанному выше алгоритму и получить $\{h'_j\}_{l=j}^{M + N \cdot K}$ элементов данной цепи. Затем необходимо отсеять первые $M$ элементов и из оставшихся выбрать каждый $N$-й элемент. Полученное множество будет являться аппроксимацией искомой выборки $\{h'_j\}_{j=1}^{N}$ \cite{russian_monte_carlo_methods}, а так как любая история выполнений $h'$ однозначно определяет значение всех случайных величин в вероятностной программе, то и аппроксимацией выборки из $P(T\ |\ X = \hat{x})$.

$K$ выбирается больше единицы, чтобы исключить автокорреляцию $h'_{i}$, которая естественным образом возникает в цепи Маркова. $M$ выбирается достаточно больш\'{и}м, чтобы независимо от выбора начальной точки $h'_1$ цепь успела «забыть» данный первоначальный выбор; это особенно важно, когда начальное априорное распределение $P(T, X)$ очень сильно отличается от апостериорного $P(T\ |\ X = \hat{x})$. Какого-то общего правила для выбора данных величин нет, и обычно они выбираются эмпирически.

\subsubsection{Программная реализация статистического вывода}
\label{sec:software_engineering_implementation_of_inference}

В предыдущем подпункте теоретически был описан алгоритм для обобщенного статистического вывода в вероятностных языках программирования.

Простая программная реализация впервые достаточно подробно была описана в \cite{wingate2011lightweight}:
\begin{enumerate}[leftmargin=1.75cm]
\item Для инициализации $h'_1$ выбирается случайным образом путем выполнения вероятностной программы и фиксации наблюдаемых случайных величин в соответствии с имеющимися данными. При этом в памяти компьютера сохраняется база данных активных случайных величин вместе с их значениями, а также сохраняется $P(h'_1)$, которое подсчитывается во время первоначального выполнения программы с учетом вероятности наблюдаемых случайных величин принять то значение, которое они приняли. В \cite{wingate2011lightweight} описывается схема адресации, которая позволяет различать случайные величины между собой.
\item Затем для выбора последующего $h'_{t+1}$ каждый раз из базы данных случайным равномерным образом выбирается случайная величина $r$ и она случайным образом варьируется в соответствии со своим локальным распределением предлагаемых переходов $\kappa(\ \cdot\ |\ r = \hat{r}, h'_t)$, которое предварительно задается для всех используемых примитивных (несоставных) случайных величин. Создается копия базы данных, в которой значение случайной величины $r$ заменяется на новое.
\item Вероятностная программа выполняется еще один раз, при этом случайные величины, для которых уже в базе данных имеется запись (согласно схеме адресации), принимают соответствующие старые значения, а $r$ принимает свое новое полученное значение.
\item В общем случае, если $r$ влияет на поток выполнения программы, генерируются значения для новых случайных величин, которые становятся активными. Для этих случайных величин прежде в базе данных не было записей.
\item Также если $r$ влияет на поток выполнения программы, то некоторые случайные величины могут перестать быть активными. В данном случае соответствующие им записи в базе данных будут невостребованы.
\item После выполнения программы второй раз можно считать, что мы получили $h'^{*}$. Также отметим, что при выполнении программы $h'^{*}$ мы получили и использовали всё необходимые компоненты для подсчета $P(h'^{*})$ и $Q(h'^{*}\ |\ h'_t)$.
\item Вероятность обратного перехода $Q(h'_t\ |\ h'^{*})$ может быть получена разными способами. Если в вероятностной программе есть только независимые случайные величины (и нет перестановочных), то первые две компоненты $Q(h'_t\ |\ h'^{*})$ могут быть найдены тривиально, так как мы знаем количество случайных величин в $h'^{*}$ и можем посчитать $\kappa(\hat{r}\ |\ r = r^{*}, h'^{*})$. Для третьей компоненты нам нужно посчитать произведение вероятностей случайных величин, которые перестали быть активными в $h'^{*}$, то есть все невостребованные случайные величины в базе данных. В случае наличия и использования перестановочных случайных величин мы можем имитировать ситуацию, что $h'^{*}$ является нашим старым состоянием, а новое состояние мы получаем в точном соответствии с $h'_t$ \cite{venture}.
\item Имея $P(h'_t)$, $P(h'^{*})$, $Q(h'^{*}\ |\ h'_t)$ и $Q(h'_t\ |\ h'^{*})$, мы подсчитываем коэффициент принятия $\alpha$ для алгоритма Метрополиса-Гастингса, и либо принимаем $h'^{*}$ с вероятностью $\alpha$, что означает, что в следующий раз мы будем использовать уже новую базу данных случайных величин со значением, соответствующими $h'^{*}$, либо отклоняем с вероятностью $(1 - \alpha)$, что означает, что на следующем шаге $h'_{t+1}$ мы снова будем использовать старую базы данную от $h'_t$.
\item Алгоритм повторяется с шага № 2.
\end{enumerate}

\subsection{Эффективность вывода}

При использовании вероятностных языков программирования встает вопрос об эффективности статистического вывода, другими словами о том, как быстро мы генерируем выборку желаемой точности из апостериорного распределения. Описанная в секции~\ref{sec:software_engineering_implementation_of_inference} программная реализация статистического вывода является неэффективной, так как происходит перевыполнение всей вероятностной программы, хотя вариация случайной величины $r$ обычно имеет только локальный эффект.

Рассмотрим более подробно данную проблему на примере следующей вероятностной программы:

\begin{lstlisting}{}{program:abc}
[ASSUME rainy-season (bernoulli 0.2)]
[ASSUME cloudy
  (bernoulli (if rainy-season 0.8 0.3))]
[ASSUME rain
  (bernoulli (if cloudy 0.8 0.2))]
[ASSUME sprinkler
  (bernoulli (if cloudy 0.1 0.5))]
[ASSUME wet-grass
  (bernoulli
    (if sprinkler (if rain 0.99 0.9)
                  (if rain 0.9 0.01)))]
\end{lstlisting}

\begin{figure}
\center
\includegraphics{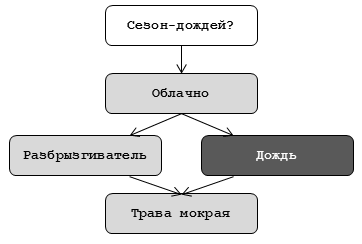}
\caption{Пример Байесовской сети. Из \cite{perov2012}.}
\label{fig:Sprinkler_Bayes_Net}
\end{figure}

Данная вероятностная программа описывает статистически простую упрощенную модель зависимости между сезоном, облачностью, дождем, работой разбрызгивателя и состоянием травы (мокрая или нет) в какой-то день. Значение переменных следующее: \texttt{rainy-season}: входит ли тот день в сезон дождей или нет?, \texttt{cloudy}: облачно в тот день или нет?; \texttt{rain}: был ли дождь в тот день?; \texttt{sprinkler}: работал ли разбрызгиватель в тот день или нет?; \texttt{wet-grass}: была ли трава мокрой в тот день? Данная модель может быть представлена с помощью Байесовской сети доверия (см. рис.~\ref{fig:Sprinkler_Bayes_Net}) и таблицами условных вероятностей. Отметим, что любая Байесовская сеть доверия может быть представлена в виде вероятностной программы на языке Church/Venture/Anglican, но не любая вероятностная программа может быть представлена Байесовской сетью.

\begin{figure}
\center
\includegraphics[width=0.5\textwidth]{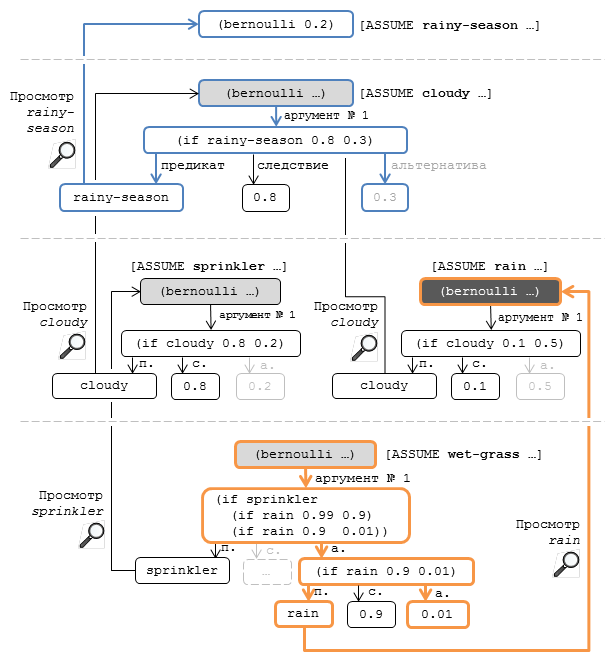}
\caption{Байесовская сеть в виде <<отпечатка>> выполнения вероятностной программы. Из \cite{perov2012}.}
\label{fig:Sprinkler_Trace}
\end{figure}

В случае, если во время очередной итерации алгоритма Метрополиса-Гастингса в качестве варьируемой случайной величины выбрана случайная величина \texttt{rain} (соответствующий узел на рис.~\ref{fig:Sprinkler_Bayes_Net} выделен самым темным цветом), для подсчета коэффициента принятия достаточно лишь рассмотреть значения и вероятности при данных значениях узлов <<Дождь>> и <<Трава мокрая>>. Все остальные значения и их вероятности останутся прежними. Иллюстрацию распространения возмущений в связи с вариацией случайной величины <<Дождь>> см. на рис.~\ref{fig:Sprinkler_Trace}.

В подобном простом примере этот факт не играет большой роли, так как отношение случайных величин, требующих <<переучета>>, к общему количеству активных случайных величин невелико, однако при большом объеме данных это играет решающую роль при выполнении статистического вывода во многих моделях при помощи программных реализаций вероятностных языков программирования. Например, в скрытых марковских моделях \cite{nikolenko_probabilistic_learning,goodman2008church,wood2014new} или в латентном размещении Дирихле \cite{blei2003latent,perov2012}.

Например, в простой скрытой Марковской модели есть $N$ скрытых и $N$ наблюдаемых величин. При реализации статистического вывода алгоритмом, описанным в секции~\ref{sec:software_engineering_implementation_of_inference}, одна итерация алгоритма Метрополиса-Гастингса имеет сложность (по времени) $O(N)$, хотя желаемая и возможная сложность $O(1)$ или по крайней мере $O(\log N)$.

Для достижения желаемой сложности в каждой истории выполнений необходимо отслеживать зависимости между случайными величинами. Следует отметить, что эти зависимости в общем случае могут изменяться в вероятностных программах. Например, в следующей программе
\begin{lstlisting}{}{program:abc}
[ASSUME a (bernoulli 0.5)]
[ASSUME b (gamma 1 1)]
[ASSUME c (normal (if a b 3.0) 1)]
\end{lstlisting}
случайная величина \texttt{c}, т.о. \texttt{(normal \ldots)}, зависит от случайной величины \texttt{b}, т.о. \texttt{(gamma \ldots)}, не всегда, а только если случайная величина \texttt{a} принимает значение <<ИСТИНА>>.

Описание структур данных и алгоритмов, необходимых для отслеживания зависимостей в режиме реального времени, были предварительно приведено в \cite{perov2012} и \cite{wu2013reduced}, а затем более обширно и подробно в \cite{venture}. При использовании данных структур данных и алгоритмов временная сложность одной итерации алгоритма Метрополиса-Гастингса в простой скрытой Марковской модели равна $O(\log N)$, при этом если использовать упорядоченный перебор случайных величин, то временная сложность снизиться до $O(1)$, так как логарифмический фактор появляется в связи с необходимостью выбирать случайным образом следующий узел (т.е. случайную величину) для вариации.

В простой скрытой Марковской модели количество <<активных>> случайных величин постоянно и вид зависимости тривиален (от каждой скрытой случайной величины напрямую зависит только одна наблюдаемая случайная величина и только следующая скрытая случайная величина). В более сложных моделях виды зависимостей более изощрены и количество случайных величин меняется от одной истории выполнений к другой. С другой стороны, для большого количества порождающих моделей, используемых в настоящее время в машинном обучении, необходимо, чтобы время на одну итерацию оставалось постоянным или росло хотя бы логарифмически вместе с линейным ростом количества наблюдений, иначе статистический вывод будет невозможен за разумное время. 

\begin{figure}
\center
\includegraphics{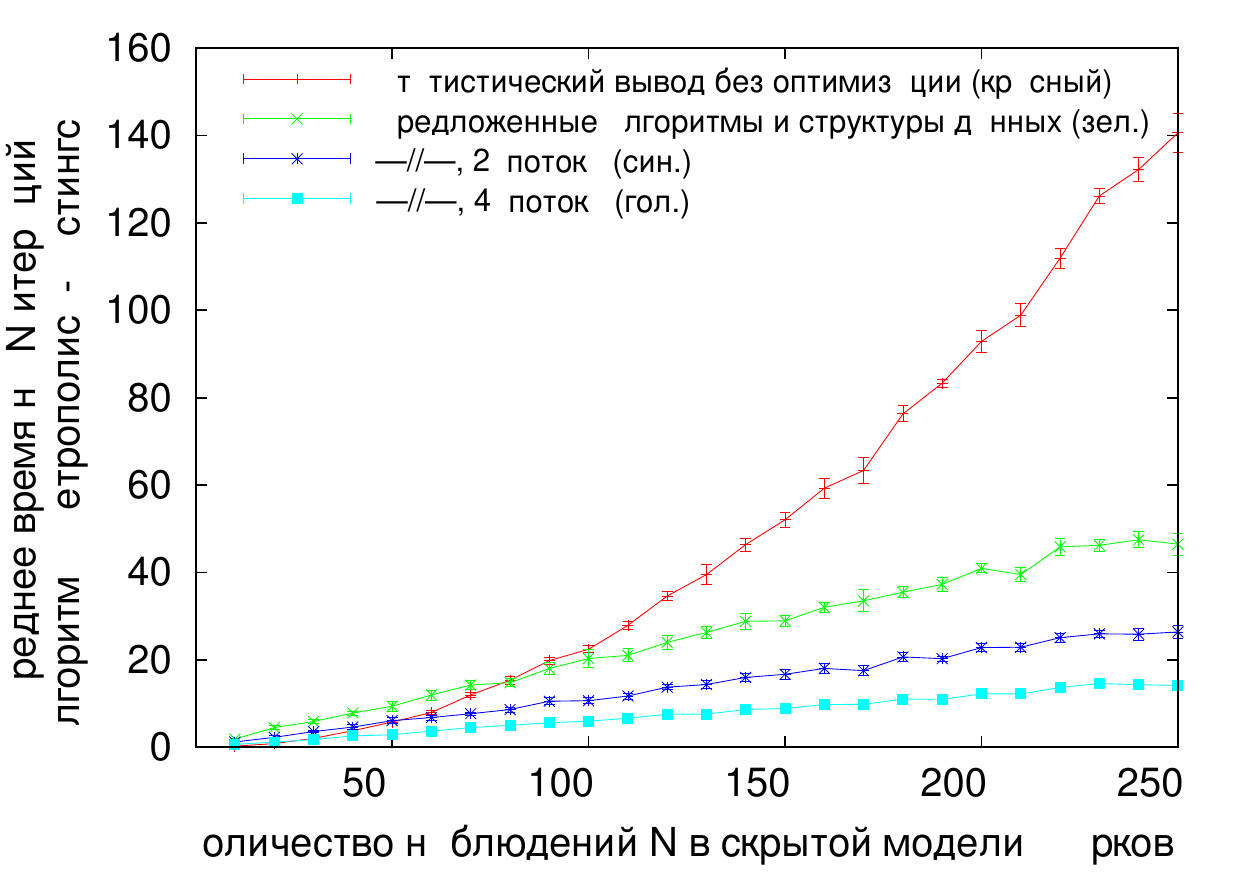}
\caption{Эффективность статистической вывода при его различных реализациях: красный график соответствует простейшему алгоритму, описанному в~\ref{sec:software_engineering_implementation_of_inference}, при котором каждый раз происходит перевыполнение всей вероятностной программы. Зеленый график соответствует использованию новых алгоритмов и структур данных, что позволяет производить статистический вывод асимптотически более эффективно. Синий и голубой график показывают, что применение предлагаемого подхода также позволяет производить статистический вывод параллельно, по крайней мере приближенно. Рис. из \cite{perov2012}.}
\label{fig:HMM_efficiency}
\end{figure}

На рис.~\ref{fig:HMM_efficiency} показаны результаты применения алгоритмов и структур данных, описанных в \cite{perov2012,wu2013reduced,venture}. При использовании старого подхода, описанного в \cite{wingate2011lightweight} (см. секцию~\ref{sec:software_engineering_implementation_of_inference}) время на $N$ итераций алгоритма Метрополиса-Гастингса росло квадратично с линейным ростом размерности модели ($N$ скрытых и $N$ наблюдаемых случайных величин), а при использовании предлагаемого подхода время растет квазилинейно.

На этом же рисунке видно, что благодаря локализации выполнения одной итерации Метрополиса-Гастингса стало возможным проводить приближенный статистический вывод параллельно.

Следует отметить, что пространственная сложность алгоритма (объем памяти, необходимый для его применения) равна, грубо говоря, $O(N + K)$, где $O(K)$ — стоимость хранения в памяти зависимостей между случайными величинами. В общем случае эта величина может быть достаточно большой, и в работе \cite{wu2013reduced} описаны возможности более эффективного расходования памяти.

\subsection{Порождающее вероятностное программирование в распознавании образов}

Как было отмечено в начале, цель вероятностного программирования --- облегчить задачу моделирования порождающих вероятностных моделей и проведения вывода в них. Примером иллюстрации успешного предварительного применения вероятностного программирования может служить работа \cite{mansinghka2013approximate}, в которой вероятностное программирование используется для моделирования вероятностной модели находящихся на изображении объектов и их взаимодействия между собой. Байесовский подход к интерпретации изображений путем задания априорного распределения на расположение объектов и на связи между ними был предложен задолго до появления рассматриваемых языков программирования, однако именно с их появлением исследование и осуществление данного подхода стало проще, так как вероятностные языки программирования позволяют композиционно и компактно представлять вероятностные порождающие модели и проводить статистический вывод в них.

В работе рассматриваются два примера: проблема графической CAPTCHA \cite{von2003captcha} --- <<компьютерного теста, используемого для определения, кем является пользователь системы: человеком или компьютером>>, знакомого почти что каждому пользователю интернета; и проблема нахождения на изображении с камеры автомобиля дороги, разделительной полосы, и левого и правого оврагов. Полученные результаты по своей эффективности на рассматриваемых простых примерах не уступают другим современным подходам к решению этих задач, однако представление, моделирование и вывод проще осуществляется с помощью вероятностного программирования.

\subsection{О различиях между Church, Venture, Anglican}
\label{sec:differences_between_languages}

Целью данной главы, очевидно, не было подробное описание этих вероятностных языков, а только краткое введение в них. Заинтересованному читателю мы можем порекомендовать продолжить свое знакомство с вероятностным языком Church с \cite{goodman2008church,probmods,ForestDB,wingate2011lightweight}, Venture --- \cite{venture}, Anglican --- \cite{tutorialIceland2014,wood2014new,Paige-ICML-2014}.

Хотя многое объединяет эти вероятностные языки, в некоторых принципиальных вещах они различаются. Например, Venture на данный момент больше позиционируется как универсальная платформа, включающая в себя разные виды алгоритмов и методов статистического вывода с запроектированной возможностью добавлять новые с помощью использования базовых компонент и методов. В рамках работы над Anglican развиваются методы обобщенного вывода с использованием методов фильтрации частиц. Church, с другой стороны, позволяет производить статистический запрос внутри другого статистического запроса, что Venture и Anglican пока делать не могут.

\clearpage

\section{Автоматическая генерация вероятностных программ}

При использовании полных по Тьюрингу вероятностных языков программирования, включающих в себя функции высших порядков\footnote{Функциями (процедурами) высших порядков называются функции, аргументами или значениями которых могут быть другие функции.}, которыми в том числе являются языки Church, Venture и Anglican, вероятностная программа одновременно является и порождающей моделью, и записанной процедурой для генерации элементов выборки из этой модели путем выполнения исходного кода данной процедуры. Любая процедура в вероятностном программировании является формально программным кодом, который описывает процесс генерации элемента выборки при заданных аргументах данной функции. Таким образом, процедуры вероятностных программ являются конструктивным способом описания условных распределений.

Полные по Тьюрингу и допускающие функции высших порядков вероятностные языки программирования открывают возможность проведения вывода исходного текста самих вероятностных программ, если задано априорное распределение на множестве исходного текста, с помощью операторов $\mathrm{eval}$ и $\mathrm{apply}$. Грубо говоря, необходимо представить вероятностную порождающую мета-модель, которая будет генерировать вероятностные модели в виде исходного кода вероятностных программ.

Данная глава основана на работе \cite{learning_probabilistic_programs}, которая включает в себя первые предварительные результаты по этой амбициозной задаче вывода самих порождающих вероятностных моделей при наличии какой-либо информации об искомом распределении, которое определяется искомой вероятностной программой. Отметим, что статистический вывод в пространстве исходного кода сложен и нет какого-то простого подхода как к построению вероятностных порождающих моделей исходного кода \cite{maddison2014structured}, так и выводу в них \cite{liang2010learning}.

В рамках нашей предварительной работы мы поставили задачу найти с помощью статистического вывода и предлагаемой нами порождающей вероятностной мета-модели такие вероятностные программы, которые будут генерировать выборку элементов, схожую по своим статистическим характеристикам с каким-то заранее заданным распределением.

Эта задача интересна сама по себе, так как нахождение эффективных алгоритмов генерации (моделирования) случайных величин --- нетривиальная задача и для людей-ученых, которой они занимаются на протяжении десятков лет \cite{marsaglia1964convenient,box1958note,Devroye86non-uniformrandom}.

Наши предварительные результаты показывают, что подобный автоматизированный вывод порождающих вероятностных моделей в виде вероятностных программ действительно возможен. В частности, мы приводим результаты успешного эксперимента, в рамках которого мы автоматически нашли обобщенную вероятностную программу, которая подлинно (не приближенно) генерирует случайные величины, распределенные по закону Бернулли с произвольным параметром $p$.

\subsection{Обзор литературы}

Рассматриваемые нами идеи относятся к разным областям, в том числе к автоматизации процесса программирования \cite{gulwani_et_al:DR:2014:4507,looks2007program}, индуктивному программированию \cite{gulwani_et_al:DR:2014:4507,muggleton1994inductive,muggleton1992golem,de2008probabilistic,kersting2005inductive,muggleton1996stochastic}, автоматическому моделированию \cite{grosse2012exploiting,duvenaud2013structure}, компьютерному определению и представлению плотности распределений. Подходы к решению проблемы включают статистический вывод, поиск и оптимизацию, в том числе эволюционные алгоритмы и в особенности генетическое программирование.

Идеи и методы использования вероятностного программирования для изучения и автоматизированного представления вероятностных моделей предлагались и ранее \cite{goodman2008church,hwang2011inducing,andy_gordon_automatic_modelling,venture}.

Насколько автору известно, описываемый подход к порождению вероятностных программам мета-вероятностной программой ранее не рассматривался в качестве отдельной проблемы достаточно подробно, хотя первые шаги в исследовании и формулировке проблемы были сделаны в работах \cite{hwang2011inducing,venture}.

\subsection{Описание подхода}

Наш подход может быть описан в рамках приближенных Байесовских вычислений \cite{marjoram2003markov} с использованием метода Монте-Карло по схеме цепей Маркова с выбором в качестве искомого апостериорного распределения
\begin{equation}
\pi(\mathcal{X}|\hat{\mathcal{X}})p(\hat{\mathcal{X}}|\mathcal{T})p(\mathcal{T}),
\label{eqn:simpleabcformulation}
\end{equation} 
где $\pi(\mathcal{X}|\hat{\mathcal{X}})$ --- вероятностная мера, измеряющая расстояние между значения статистик, вычисленных соответственно на выборке $\mathcal{X}$ из искомого распределения и выборке $\hat{\mathcal{X}}$, полученной путем выполнения исходного кода вероятностной программы-кандидата.

Пусть существует распределение $F$ с параметрами $\theta$, вероятностную программу для генерации элементов выборки из которого мы хотим вывести. Пусть $\mathcal{X} = \{x_i\}_{i=1}^I, x_i \sim F(\cdot | \theta)$ будет выборкой из $I$ элементов из распределения $F$ при каком-то фиксированном значении параметров $\theta$. Рассмотрим задачу вывода исходного кода вероятностной программы $\mathcal{T}$, которая при ее неоднократном выполнении $J$ раз сгенерирует выборку элементов $\hat{\mathcal{X}}= \{\hat{x_j}\}_{j=1}^J$, $\hat{x_j} \sim \mathcal{T}(\cdot)$, статистически близких к распределению $F$ при заданных параметрах $\theta$.

Пусть $s$ будет статистикой выборки, и тогда $s(\mathcal{X})$ и $s(\hat{\mathcal{X}})$ --- значение этой статистики на элементах выборки из $F$ и из $\mathcal{T}(\cdot)$ соответственно. Пусть вероятностная мера $d(s(\mathcal{X}),s(\hat{\mathcal{X}})) = \pi(\mathcal{X}|\hat{\mathcal{X}})$, для простоты ненормированная, принимает б\'{о}льшие значения, когда $s(\mathcal{X}) \approx s(\hat{\mathcal{X}})$. $d$ можно интерпретировать как расстояние или штраф.

Мы используем вероятностное программирование для представления мета-модели, порождающей другие вероятностные программы в виде исходного текста, и для проведения статистического вывода в пространстве искомых вероятностных моделей. Для статистического вывода мы использовали программную реализацию вероятностного языка программирования Anglican, которая поддерживает \cite{wood2014new} статистический вывод методом частиц и методом Метрополиса-Гастингса по методу Монте-Карло по схеме цепей Маркова.

Вероятностная мета-модель представлена на рис.~\ref{fig:inferring_std_normal_via_moments}, где на первой строке мы устанавливаем соответствие между $\mathcal{T}$ и переменной \texttt{program-text}, которая будет содержать один сгенерированный элемент из распределения на исходный код $P(\mathcal{T})$, определенное априорно через порождающую процедуру \texttt{production} с помощью адаптивной грамматики по типу \cite{johnson2007adaptor} (см. подробнее в разделе~\ref{sec:grammar}).

\begin{figure}
\lstset{
  basicstyle=\large,
  numbers=none,
  language=Lisp
}
\lstset{language=Lisp}
\begin{lstlisting}
[ASSUME program-text (productions `() `real)]
[ASSUME program (eval (list `lambda `() program-text))]
[ASSUME samples (apply-n-times program 10000 '())]
[OBSERVE (normal (mean samples) noise-level) 0.0]
[OBSERVE (normal (variance samples) noise-level) 1.0]
[OBSERVE (normal (skewness samples) noise-level) 0.0]
[OBSERVE (normal (kurtosis samples) noise-level) 0.0]
[PREDICT program-text]
[PREDICT (apply-n-times ... (program))]
\end{lstlisting}
\vspace{-4mm}
\caption{Вероятностная программа для вывода исходного кода вероятностной программы для генерации случайных чисел, распределенных по закону стандартного нормального распределения $\mathrm{Normal}(0, 1)$.}
\label{fig:inferring_std_normal_via_moments}
\end{figure}

Мы не указываем здесь $\theta$, так как задача вывода в данном случае найти вероятностную программу, генерирующую элементы из {\it стандартного} нормального распределения. Переменная \texttt{samples} на второй строке представляет описанную выше выборку $\hat{\mathcal{X}}$ из вероятностной программы-кандидата, и в этом примере $J=10000$.

$s$ и $d$ вычисляются на следующих четырех строках вероятностной программы, где статистика $s$ определяется как четырехмерный вектор, включающий в себя соответственно выборочные среднее, дисперсию, коэффициент асимметрии и коэффициент эксцесса выборки элементов из распределения, определенного вероятностной программой, полученной из распределения $\mathcal{T}$. Мера расстояния $d$ определяется через плотность многомерного нормального распределения со средними $[0.0, 1.0, 0.0, 0.0]^T$ и диагональной ковариационной матрицей $\sigma^2\mathbf{I}$. Отметим, что это означает, что мы ищем вероятностные программы, результат выполнения которых определяет распределение со средним равным 0, дисперсией --- 1, коэффициентами асимметрии и эксцесса равными 0, и мы «штрафуем» отклонения от этих значений с помощью квадратичной экспоненциальной функции потерь с коэффициентом $\frac{1}{\sigma^2}$, где $\sigma$ определена как \texttt{noise-level}. Эта функция потерь представляет собой функцию плотности нормального распределения.

Данный пример служит хорошей иллюстрацией основных особенностей нашего подхода. Для поиска в виде вероятностной программы генератора случайных чисел из стандартного нормального распределения мы используем аналитическую информацию $s(F)$ о стандартном нормальном распределении при вычислении расстояния между статистиками $d(s(F),s(\hat{\mathcal{X}}))$. Существует по крайней мере три различных ситуации, включая данную, в которых $s$ и $d$ могут вычисляться различными способами:
\begin{enumerate}[leftmargin=1.75cm]
\item В рамках первой ситуации мы ищем вероятностную программу, которая эффективно генерирует элементы из аналитически известных распределений. Под эффективностью в данном случае можно понимать вычислительную временную сложность и среднее количество использованной энтропии для генерации элемента выборки в среднем. Практически всегда в этой ситуации и при подобной постановке задачи статистики распределения $F$ известны аналитически.
\item Вторая ситуация возникает, когда мы можем генерировать только элементы выборки из $F$. Например, подобная ситуация возникает, когда мы генерируем элементы из апостериорного распределения с помощью <<дорогостоящего>> (вычислительно) метода Монте-Карло по схеме Марковских цепей, и мы заинтересованы получить другое представление апостериорного распределения в виде вероятностной программы, чье априорное распределение будет точно или хотя бы приблизительно совпадать с искомым апостериорным.
\item При третьей ситуации нам заранее дана фиксированного размера выборка из $F$, и мы хотим найти вероятностную программу, которая позволит генерировать выборку произвольного размера из $F$ в дальнейшем. Мы начали этот раздел с постановки задачи именно в рамках третьей ситуации.
\end{enumerate}

Следует отметить, что возможная польза представления генератора случайных чисел потенциально не только в том, чтобы эффективно и быстро генерировать выборку из желаемого распределения, но и чтобы получить формальное представление желаемого распределения или его приближения в виде исходного кода вероятностной программы, то есть в виде формальной сущности, дальнейшие действия с которой можно производить; в том числе проводить анализ над выведенным исходным кодом и использовать найденные блоки исходного кода для решения других задач.

Рис.~\ref{fig:inferring_bernoulli_via_gtest} иллюстрирует другое важное обобщение решение задачи, сформулированной в самом начале с апостериорным распределением \eqref{eqn:simpleabcformulation}.

\begin{figure}
\lstset{
  basicstyle=\large,
  numbers=none,
  language=Lisp
}
\lstset{language=Lisp}
\begin{lstlisting}
[ASSUME program-text (productions `(real) `bool)]
[ASSUME program (eval (list `lambda `() program-text))]
[ASSUME J 100]
[ASSUME samples-1 (apply-n-times program J '(0.5))]
[OBSERVE (flip (G-test-p-value
                  samples-1 `Bernoulli (list 0.5))) true]
 ...
[ASSUME samples-N (apply-n-times program J '(0.7))]
[OBSERVE (flip (G-test-p-value
                  samples-N `Bernoulli (list 0.7))) true]
[PREDICT program-text]
[PREDICT (apply-n-times program J '(0.3))]
\end{lstlisting}
\vspace{-4mm}
\caption{Вероятностная программа для вывода исходного кода вероятностной программы, генерирующей случайные числа, распределенные по закону Бернулли $\mathrm{Bernoulli}(\theta)$. На предпоследней строчке выводится текст вероятностной программы-кандидата. На последней строчке вероятностная программа-кандидат выполняется \texttt{J} раз для генерации выборки из $J$ элементов при параметризации $p = 0.3$, причем предыдущие $\theta_n$ (т.о. тренировочные значения параметров) не содержали $p = 0.3$.}
\label{fig:inferring_bernoulli_via_gtest}
\end{figure}

При постановке задачи на вывод генератора случайных чисел, распределенных согласно стандартному нормальному распределению, мы не параметризовали искомое распределение никаким образом, так как у стандартного нормального распределения нет параметров, т.е. $\theta = \emptyset$. В общем случае распределения, которые мы хотим представить в виде вероятностных программ, имеют нетривиальную параметризацию, и представляют собой по сути семейство распределений. Мы хотим найти вероятностную программу, входные аргументы которой как раз бы и являлись параметрами искомого распределения, таким образом, эта вероятностная программа позволяла бы генерировать случайные величины из всего семейства. Для наглядности рассмотрим алгоритм генерации случайных чисел, распределенных по закону нормального распределения, с помощью преобразования Бокса-Мюллера, представленный в виде вероятностной программы на рис.~\ref{fig:human_written_program_text}.

\begin{figure}
\lstset{
  basicstyle=\footnotesize,
  numbers=none,
  language=Lisp
}
\begin{tabular}{ m{0.42\textwidth} m{0.45\textwidth} }
\begin{lstlisting}{}{program:abc}
[ASSUME box-muller-normal
  (lambda (mean std)
    (+ mean (* std
      (* (cos (* 2 (* 3.14159
      (uniform-continuous 0.0 1.0))))
      (sqrt (* -2
        (log (uniform-continuous 0.0 1.0)
          )))))))]
\end{lstlisting} & \begin{lstlisting}{}{program:abc}
[ASSUME poisson (lambda (rate)
  (begin (define L (exp (* -1 rate)))
         (define inner-loop (lambda (k p)
         (if (< p L) (dec k)
           (begin (define u
                     (uniform-continuous 0 1))
             (inner-loop (inc k) (* p u))))))
         (inner-loop 1 (uniform-continuous 0 1))))]
\end{lstlisting} \\
\end{tabular}
\vspace{-5mm}
\caption{Найденные и записанные людьми исходные коды вероятностных программ для {\it (слева)} общего нормального распределения $\textrm{Normal}(\mu, \sigma)$ \cite{box1958note} и {\it (справа)} распределения Пуассона $\textrm{Poisson}(\lambda)$ \cite{knuth1998art}. Эти исходные коды входят в собранный нами корпус, с помощью которого мы определяем априорные вероятности для наших порождающих правил путем подсчета количества встречающихся констант, процедур разных видов и т.д.}
\label{fig:human_written_program_text}
\end{figure}

Эта вероятностная процедура параметризована двумя параметрами: средним и стандартным отклонением. Для постановки обобщенной задачи вывода параметризованных вероятностных программам нам необходимо изменить наше искомое апостериорное распределение, включив в него параметр $\theta$, параметризующий искомое распределение $F$:
\begin{equation}
\pi(\mathcal{X}|\hat{\mathcal{X}},\theta)p(\hat{\mathcal{X}}|\mathcal{T}, \theta)p(\mathcal{T}|\theta)p(\theta).
\label{eqn:completeabcformulation}
\end{equation} 

Нам бы хотелось вывести вероятностную программу, которая смогла бы обобщить все возможные значения параметра $\theta$. С допущением, что если мы выберем конечное число $N$ различных параметризаций $\hat{\theta}_i$, мы получим обобщение всего семейства распределений в виде вероятностной программы, мы формулируем нашу задачу в виде следующего приближения с использованием приближенных Байесовских вычислений в рамках метода Монте-Карло по схеме цепей Маркова:
\begin{equation}
\frac{1}{N}\sum_{n=1}^N \pi(\mathcal{X}_n|\hat{\mathcal{X}}_n,\theta_n)p(\hat{\mathcal{X}}_n|\mathcal{T}, \theta_n)p(\mathcal{T}|\theta_n) \approx \int \pi(\mathcal{X}|\hat{\mathcal{X}},\theta)p(\hat{\mathcal{X}}|\mathcal{T}, \theta)p(\mathcal{T|\theta})p(\theta) d\theta.
\label{eqn:marginalizedabcformulation}
\end{equation} 

Вероятностная программа для поиска параметризованной вероятностной программы, генерирующей случайные числа, распределенных по закону Бернулли $\mathrm{Bernoulli}(\theta)$, представлена на рис.~\ref{fig:inferring_bernoulli_via_gtest}, и наглядным образом иллюстрирует применение данного допущения и приближения. При выбранных $N$ различных параметризациях параметра $p$ распределения Бернулли мы каждый раз генерируем $J$ элементов из вероятностной программы-кандидата, аккумулируя расстояние (штраф) между искомым распределением и полученным распределением, представляющим вероятностную программу-кандидата. В каждом конкретном случае для $\theta_i$ мы высчитываем: 1) расстояние с использованием статистики G-теста (более <<современный>> аналог критерия согласия Пирсона и соответствующей статистики) в виде
\[G_n= 2 \sum_{i\in{0,1}} \#[\hat{\mathcal{X}}_n = i]\mathrm{ln}\left(\frac{\#[\hat{\mathcal{X}}_n = i]}{\theta_n^i(1-\theta_n)^{(1-i)} \cdot |\hat{\mathcal{X}}_n|}\right),\]
где $\#[\hat{\mathcal{X}}_n = i]$ --- количество элементов выборки из $\hat{\mathcal{X}}_n$, принимающих значение $i$; 2) а также соответствующее $p$-значение с нулевой гипотезой, утверждающей, что элементы выборки $\hat{\mathcal{X}}_n$ являются и элементами выборки из распределения $\mathrm{Bernoulli}(\theta_n)$. Так как распределение статистики G-теста приблизительно распределено по закону распределения хи-квадрат, т.е.~$G\sim\chi^2(1)$ в нашем примере, мы можем представить и находить расстояние $d$ в данном случае путем вычисления вероятности ложного отклонения нулевой гипотезы $H_0 : \hat{\mathcal{X}}_n \sim \mathrm{Bernoulli}(\theta_n)$. Ложное отклонение нулевой гипотезы эквивалентно успеху в проведении испытания Бернулли с вероятностью успеха равной $p$-значению.


\subsection{Грамматика и порождающие правила}
\label{sec:grammar}
С учетом наличия в нашем распоряжении выразительного вероятностного языка программирования, допускающего функции высших порядков и полного по Тьюрингу, наше априорное распределение об исходном коде искомых вероятностных программ также достаточно выразительно. В общих чертах оно схоже с адаптивными грамматиками \cite{johnson2007adaptor}, используемыми в \cite{liang2010learning}, но имеет отличия, в частности связанные с созданием сред с локальными переменными. В виде псевдокода наше априорное распределение может быть представлено следующим образом (символ $\rightarrow$ означает <<может перейти в>>):

\begin{enumerate}[leftmargin=1.75cm]
\item Выражение $expr_{type} | env \rightarrow$ в имя переменной, случайно выбираемой из среды переменных $env$ с типом $type$.
\item Выражение $expr_{type} | env \rightarrow$ в случайную константу типа $type$. Константы различных типов (целочисленные, вещественные и т.д.) генерируются из отдельного для каждого типа $type$ процесса Дирихле\footnote{Не нужно путать с распределением Дирихле.} $DP_{type}(H_{type}, \alpha)$, где базовое распределение $H_{type}$ само по себе в общем случае являются смесью нескольких распределений. Например, для констант вещественного типа мы используем смесь нормального распределения \texttt{(normal 0 10)}, равномерного непрерывного распределения \texttt {(uniform-continuous -100 100)} и равномерного дискретного распределения из множества $\{-1, 0, 1, \ldots\}$.
\item Выражение $expr_{type} | env \rightarrow (procedure_{type}\ expr_{arg\_1\_type}\ ...\  expr_{arg\_N\_type})$, где процедура $procedure$ является случайно выбираемой примитивной детерминированной или стохастической (не составной) процедурой, определенной заранее в глобальной среде, с типом возвращаемого значения $type$.
\item Выражение $expr_{type} | env \rightarrow (compound\_procedure_{type}\ expr_{arg\_1\_type}\ ...\  expr_{arg\_N\_type})$, где {\tt $compound\_procedure_{type}$} является составной процедурой, также генерируемой в соответствии с процессом Дирихле $DP_{type}(G_{type}, \beta)$, где базовое распределение $G_{type}$ случайным образом генерирует составную процедуру с типом возвращаемого значения $type$ и количеством входных аргументов, распределенным по закону Пуассона, где каждый входной аргумент имеет свой произвольный тип. Тело $body$ составной процедуры генерируется также случайным образом согласно этим же порождающим правилам и грамматике, но с учетом введение локальной среды, включающей в себя входные аргументы процедуры.
\item Выражение $expr_{type} | env \rightarrow (let\ [(gensym)\ expr_{real} ]\ expr_{type} | env \cup gensym))$, где $ env \cup gensym$ означает среду, дополненную переменной с именем $(gensym)$ и со значением, вычисляемым в соответствии с генерируемым случайным образом выражением согласно этих же порождающих правил.
\item Выражение $expr_{type} | env \rightarrow (if\ (expr_{bool})\ expr_{type}\ expr_{type})$.
\item Выражение $expr_{type} | env \rightarrow (recur\ expr_{arg\_1\_type}\ ...\  expr_{arg\_M\_type})$, таким образом рекурсивный вызов текущей составной процедуры.
\end{enumerate}

Во избежание вычислительных ошибок во время выполнения сгенерированных процедур мы заменяем примитивные функции их <<защищенными>> аналогами, например \texttt{log(a)} заменяется на \texttt{safe-log(a)}, причем последний возвращает $0$ если $a < 0$; или например \texttt{uniform-continuous} заменяется \texttt{safe-uc(a, b)}, которая в случае если $a > b$ меняет аргументы местами, а также возвращает просто $a$ если аргументы равны: $a = b$.

Множество типов, которые мы использовали в рамках наших экспериментов, включало вещественные и булевы типы, а общее множество примитивных процедур, включенных нами в глобальную среду, включало в себя такие функции как \texttt{+, $-$, *, safe-div, safe-uc, safe-normal}.  

\subsection{Вероятности использования порождающих правил}

Для задания априорных вероятностей порождающих правил, то есть вероятностей, с которой каждое из правил будет применяться в случае возможности его применения, мы вручную составили небольшой корпус вероятностных программ, которые повторяют найденные учеными \cite{Devroye86non-uniformrandom} алгоритмы генераторов случайных чисел. Примеры таких программ представлены на рис.~\ref{fig:human_written_program_text}. Заметим, что все они требуют наличия только одной стохастической процедуры, а именно \texttt{uniform-continuous}, так что мы включили только ее в глобальную среду с положительной вероятностью для экспериментов, описанных в~\ref{sec:classical_distribution}.

Используя данный корпус, мы вычислили априорные вероятности каждого порождающего правила, при этом при выводе вероятностной программы для генерации случайных величин из искомого распределения $F$ (например, распределения Бернулли), мы исключали из корпуса все элементы, которые генерируют случайные величины согласно закону распределения $F$. После этого вероятности использования порождающих правил были <<смягчены>> с помощью распределения Дирихле. В будущем можно использовать более обширные корпусы вероятностных программ, примером зарождающегося подобного корпуса может служить \cite{ForestDB}.

\subsection{Эксперименты}

Эксперименты были разработаны таким образом, чтобы проиллюстрировать все три вида возможных ситуаций, описанных ранее. Но в начале мы проиллюстрируем выразительность нашего априорного распределения исходного кода вероятностных программ. После этого мы опишем постановку и результаты экспериментов в рамках использования нашего подхода во всех трех различных ситуациях возможности вычисления расстояния $d$.

\subsubsection{Выборки из сгенерированных вероятностных программ}

\begin{figure}
\includegraphics[width=0.33\textwidth]{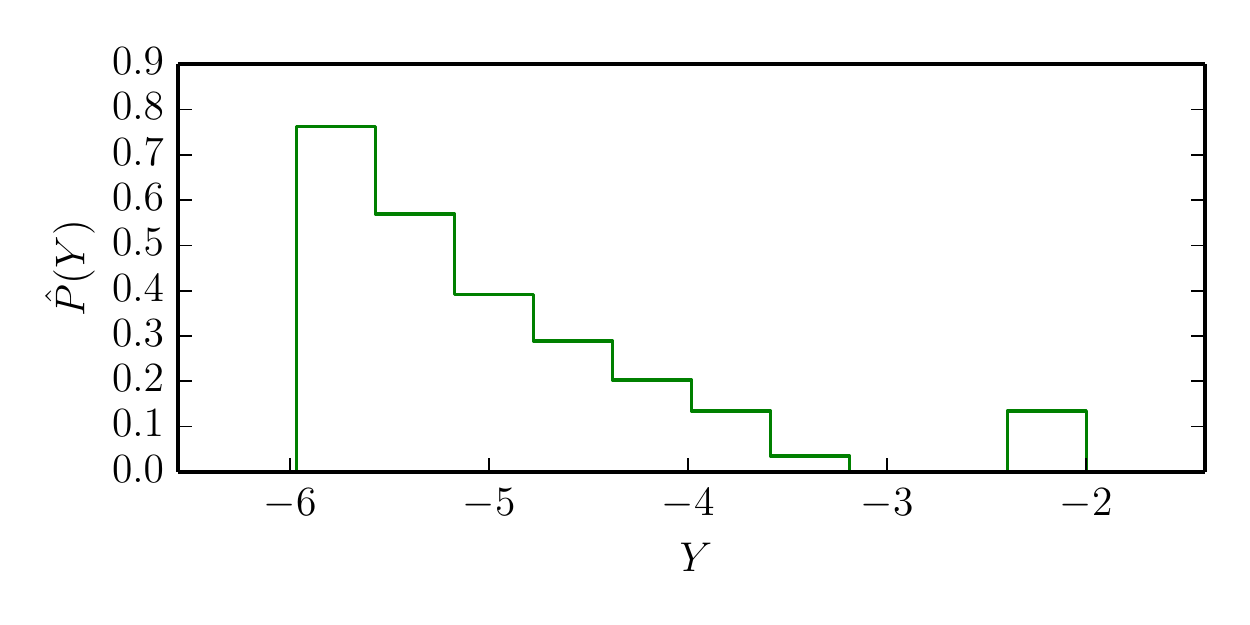}
\includegraphics[width=0.33\textwidth]{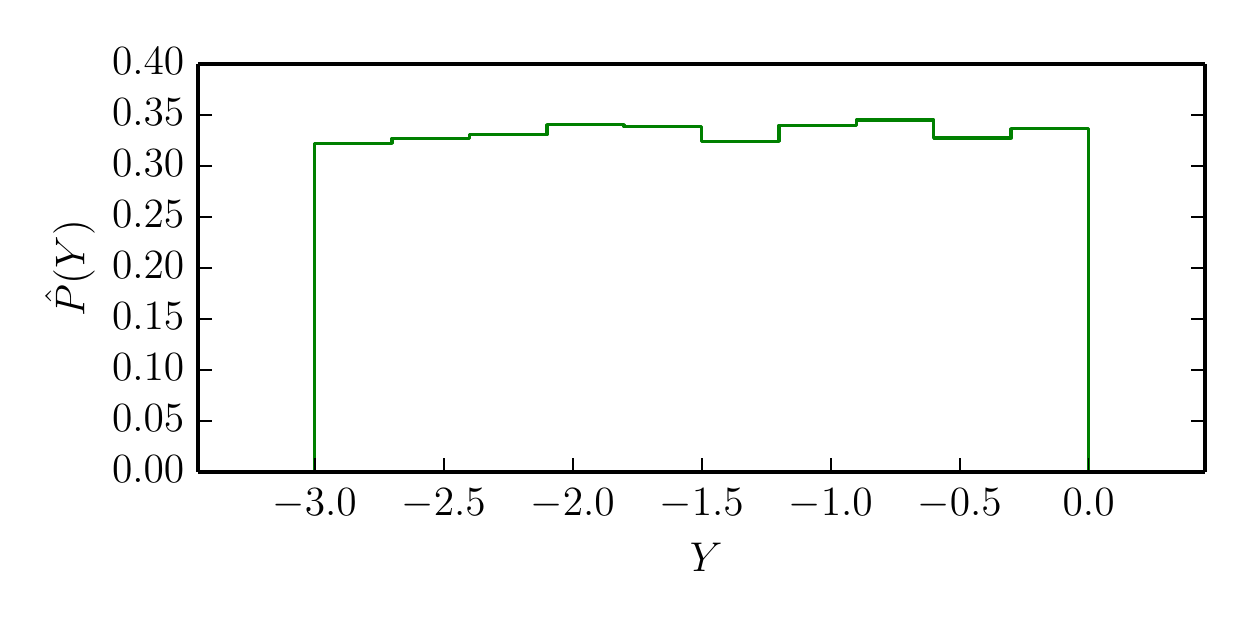}
\includegraphics[width=0.33\textwidth]{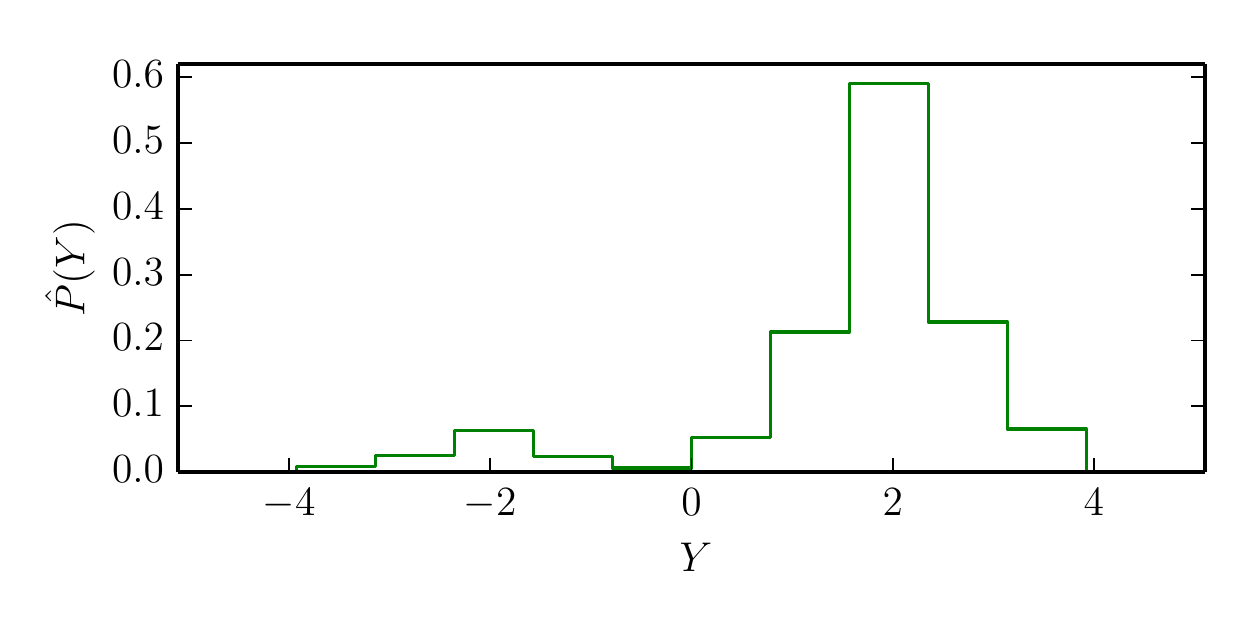}
\includegraphics[width=0.33\textwidth]{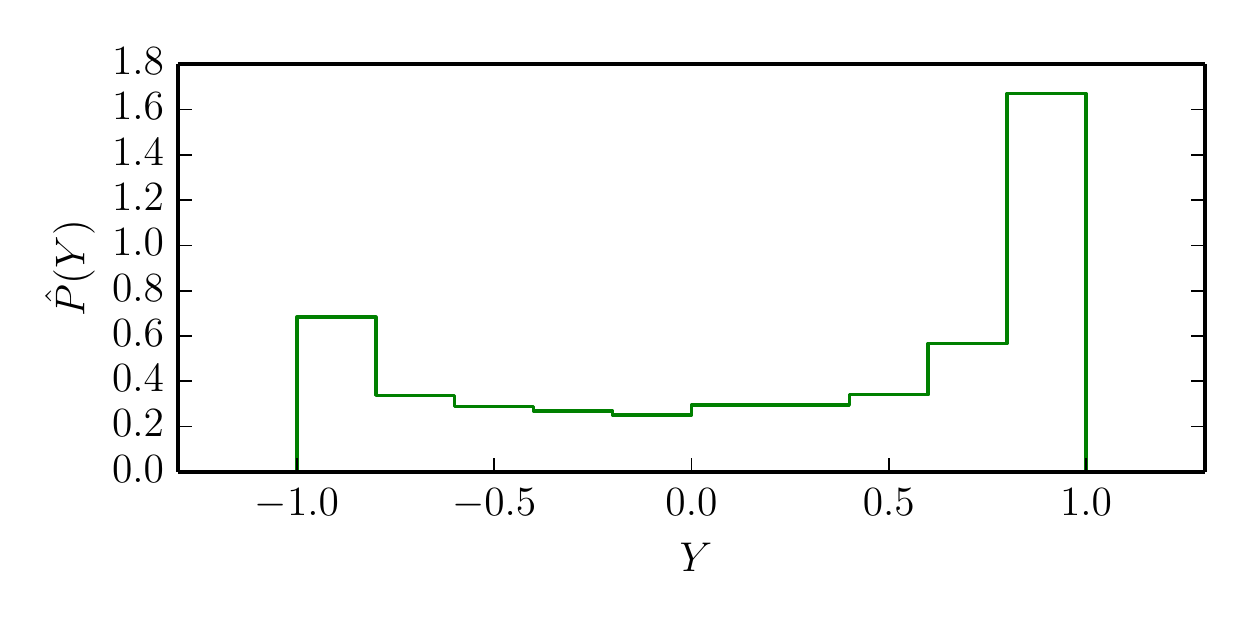}
\includegraphics[width=0.33\textwidth]{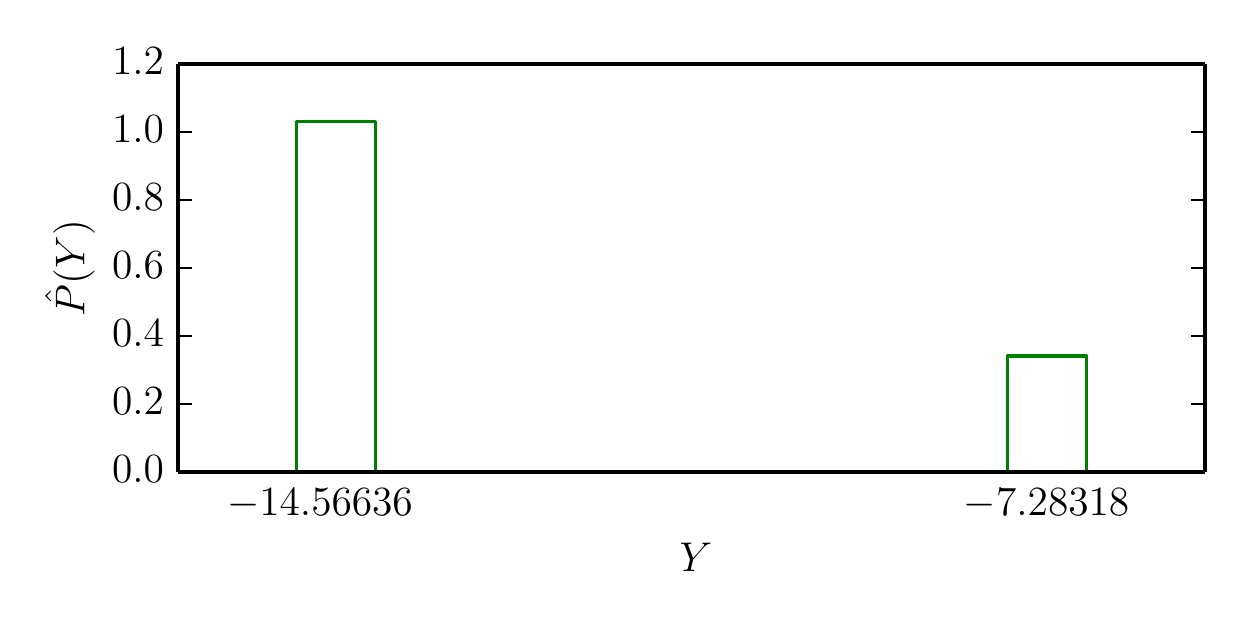}
\includegraphics[width=0.33\textwidth]{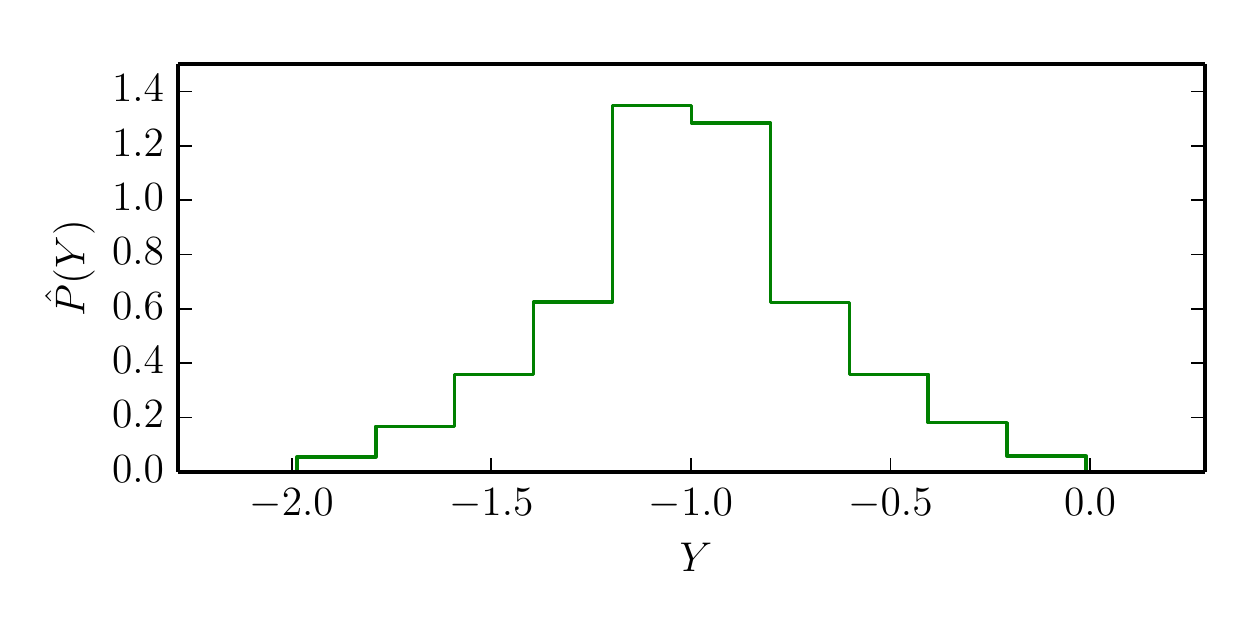}
\caption{Гистограммы выборок из некоторых порожденных вероятностных программ из их априорного распределения. По форме распределений видно, что наши порождающие правила достаточно обширны, чтобы соответствующие порождающим вероятностным программам распределения были нетривиальны.}
\label{fig:samples_from_probabilistic_programs_from_prior}
\end{figure}

Для иллюстрации выразительности нашего априорного распределения исходных кодов вероятностных программ мы приводим выборки из случайным образом сгенерированных вероятностных программ из их априорного распределения.
Эти шесть выборок в виде гистограмм из шести различных автоматически сгенерированных вероятностных программам расположены на рис.~\ref{fig:samples_from_probabilistic_programs_from_prior}.

Из рисунка видно, что разные случайным образом сгенерированные вероятностные программы определяют в общем случае достаточно различные структурно распределения. В частности, можно заметить разнообразие в носителе, дисперсии и количестве <<режимов>> (т.е. островков носителя с высокой вероятностью по отношению к очень низковероятностным пространствам между островками).

\subsubsection{Вывод вероятностных программ, определяющих распределения, совпадающие или приближающие классические одномерные распределения}
\label{sec:classical_distribution}

Как уже было отмечено, для всех классических одномерных распределений существуют алгоритмы, позволяющие и точно, и приближенно генерировать любое количество элементов из данных параметризованных распределений, и они были аналитически выведены учеными. Эти алгоритмы в том числе, очевидно, могут быть записаны как вероятностные программы в виде их исходного кода.

Мы провели серию экспериментов, чтобы проверить возможность автоматического вывода вероятностных программ, генерирующих выборки из классических одномерных распределений, а именно из распределения Бернулли $\mathrm{Bernoulli}(p)$, распределения Пуассона $\mathrm{Poisson}(\lambda)$, Гамма-распределения $\mathrm{Gamma}(a, 1.0)$, Бета-распределения $\mathrm{Beta}(a, 1)$, стандартного нормального распределения $\mathrm{Normal}(0, 1)$, и <<общего>> нормального распределения $\mathrm{Normal}(\mu, \sigma)$.

Для каждого семейства распределений мы проводили статистический вывод, выбирая в качестве целевого апостериорного распределения частное распределение $\pi(\mathcal{X}|\hat{\mathcal{X}})p(\hat{\mathcal{X}}|\mathcal{T})p(\mathcal{T})$, маргинализированное по параметру $\theta$. Для приближения в каждом случае мы выбирали малое множество значений параметров $\left\{\theta_1, \ldots, \theta_N\right\}$ и определяли апостериорное распределение ограничениями по $p-$значению (для распределения Бернулли) и по близости моментам к ожидаемым (для всех других рассматриваемых распределений). Еще раз отметим, что при задании априорного распределения на порождающие правила в каждом конкретном случае из корпуса исключались все вероятностные программы, относящиеся к искомому распределению.

\begin{figure}
\includegraphics[width=0.33\textwidth]{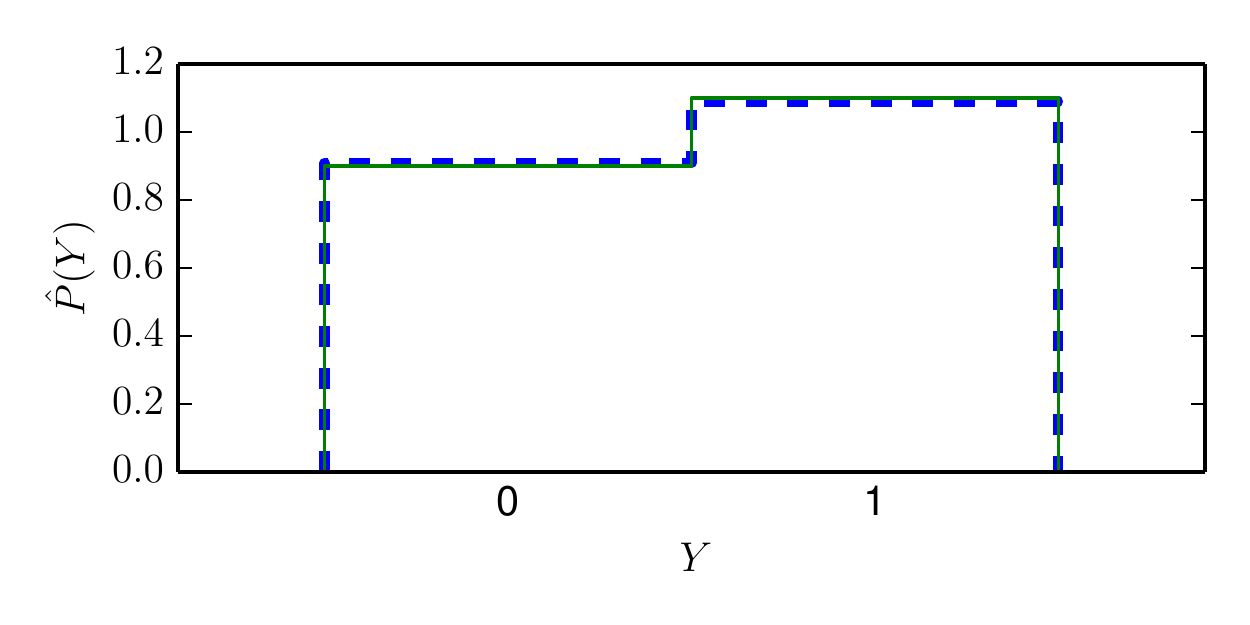}
\includegraphics[width=0.33\textwidth]{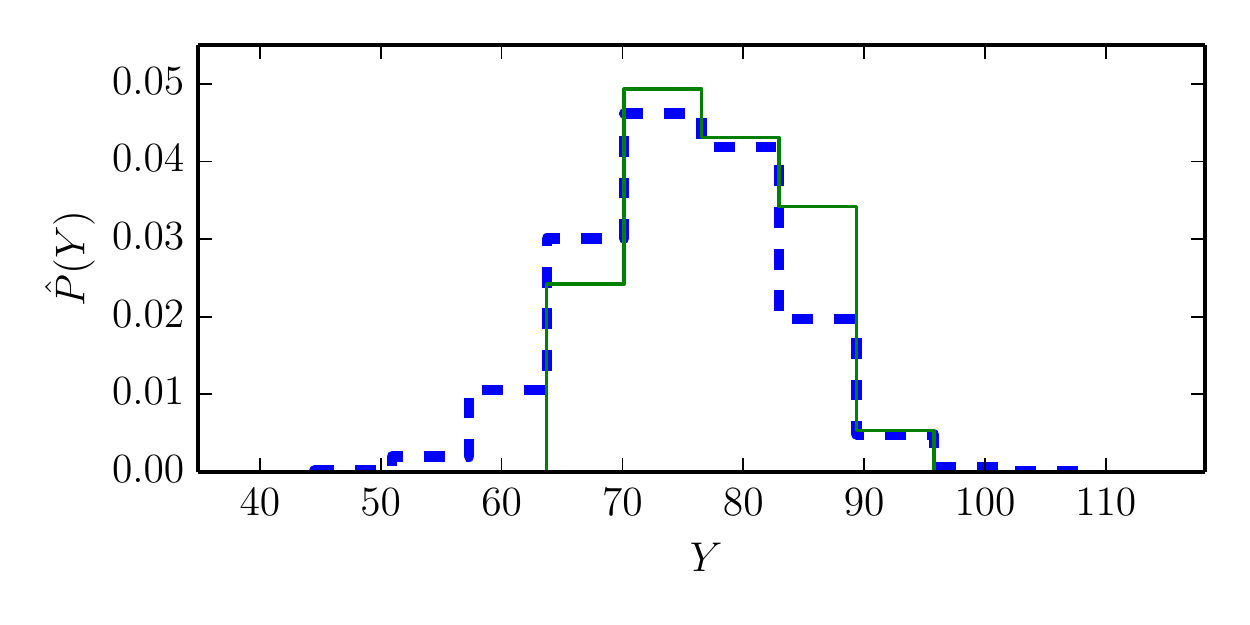}
\includegraphics[width=0.33\textwidth]{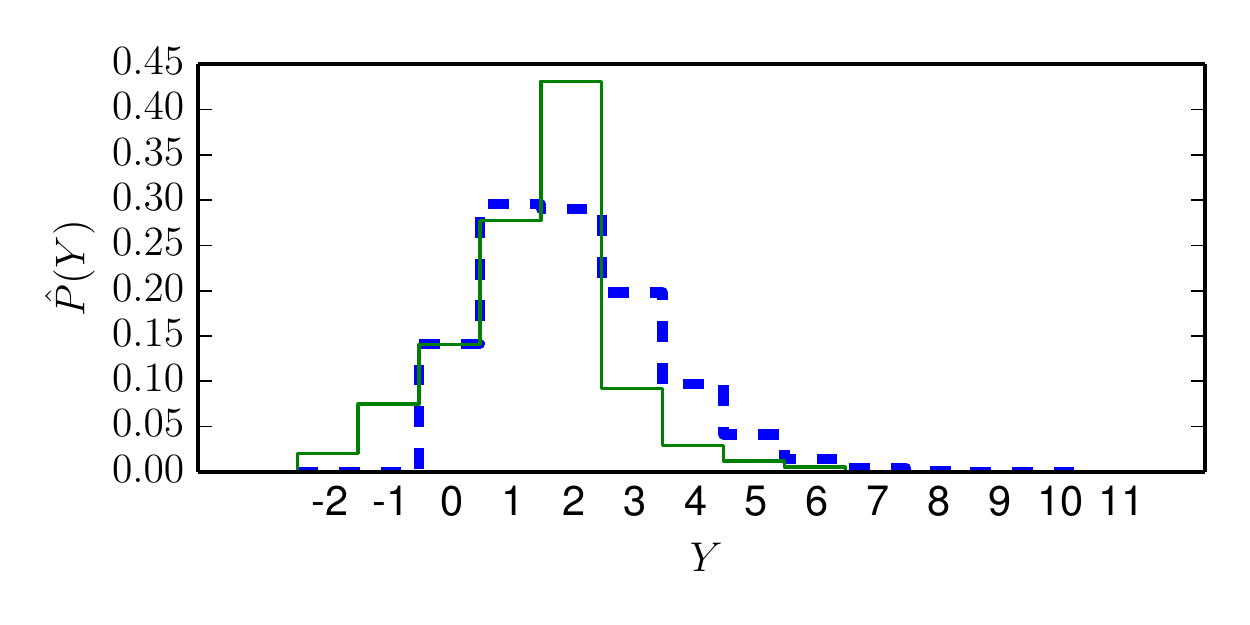}
\includegraphics[width=0.33\textwidth]{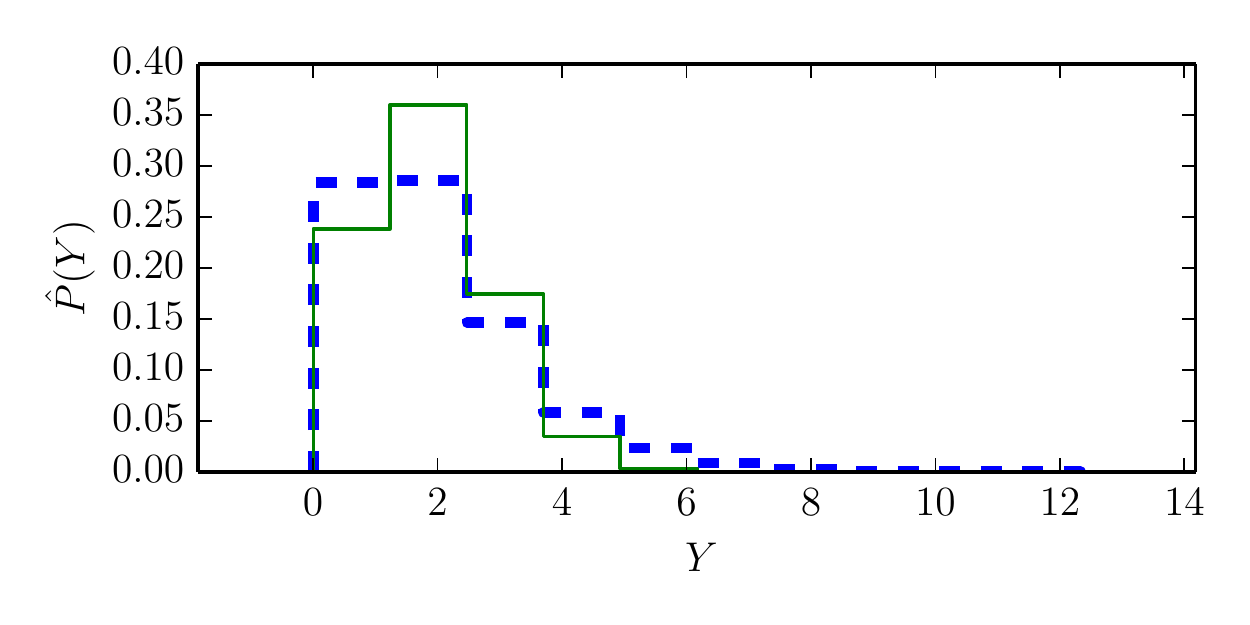}
\includegraphics[width=0.33\textwidth]{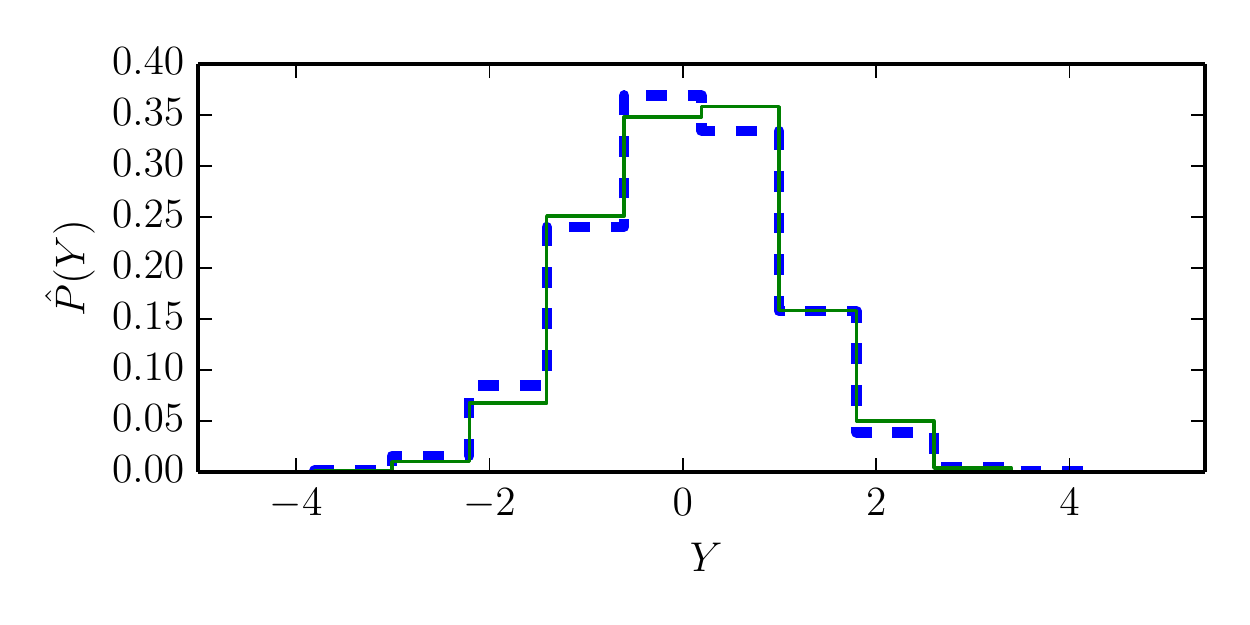}
\includegraphics[width=0.33\textwidth]{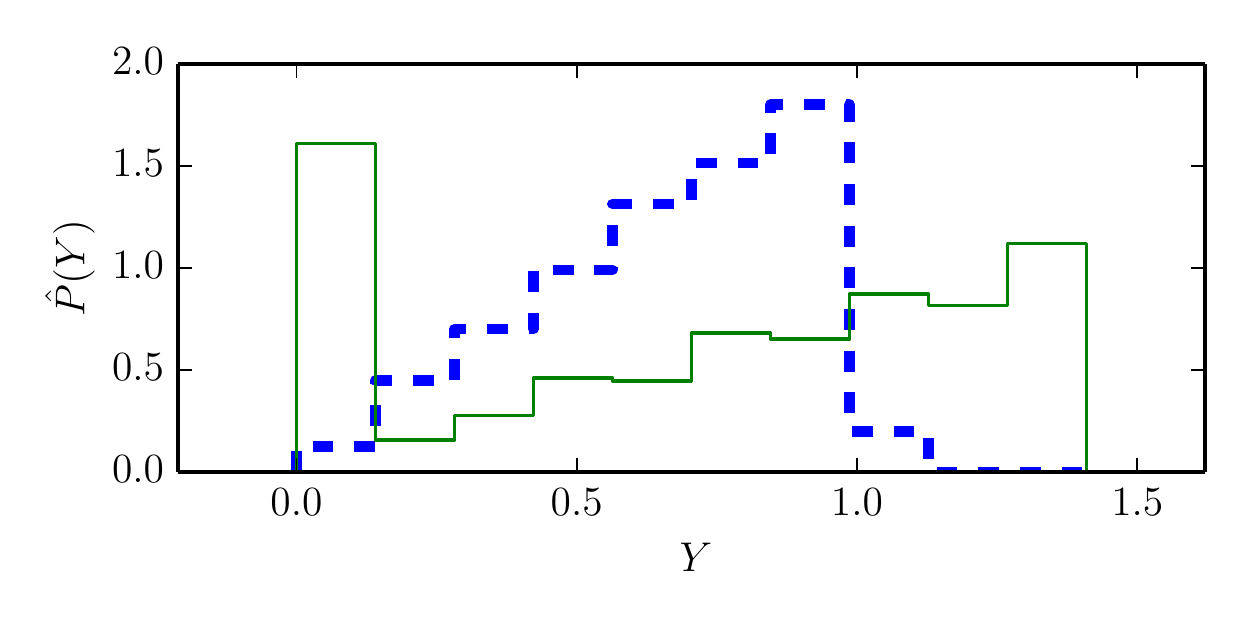}
\caption{{\it (Зеленые сплошные линии)} Гистограммы выборок из распределений, соответствующих вероятностным программам, имеющим <<высокую>> вероятность в апостериорном распределении при поиске порождающих моделей, соответствующих распределениям (слева направо, сверху вниз): $\mathrm{Bernoulli}(p)$, $\mathrm{Normal}(\mu, \sigma)$, $\mathrm{Poisson}(\lambda)$,
 $\mathrm{Gamma}(a, 1.0)$,
$\mathrm{Normal}(0, 1)$,  
$\mathrm{Beta}(a, 1.0)$. {\it (Синие пунктирные линии)} Гистограммы выборок из <<настоящих>> распределений. Параметризация семейства распределений в каждом случае производилась другими значениями параметров, которые не входили в обучающее множество значений $\{\theta_n\}$. Отметим, что для семейства распределений Бернулли $\mathrm{Bernoulli}(p)$ была выведена вероятностная программа, статистически точно генерирующая элементы выборки при любой параметризации (см. рис.~\ref{program:Bernoulli_samplers}). С другой стороны, не все распределения получили  хорошее приближение найденными вероятностными программами за то время, что мы проводили вывод, как, например, в случае Бета-распределение $\mathrm{Beta}(a, 1.0)$.
}
\label{fig:common_distributions_histograms}
\end{figure}

Образцы гистограмм выборок из лучших найденных в результате вывода вероятностных программ представлены на рис.~\ref{fig:common_distributions_histograms}, где под лучшими мы понимаем вероятностные программы с наименьшим расстоянием $d$ на тренировочных значениях параметров и соответствующих значениях моментов или максимального $p$-значения.

Стоит особенно отметить результат эксперимента с распределением Бернулли, в рамках которого был найден исходный код вероятностной программы, статистически подлинно (точно) генерирующей выборку из всего семейства распределения Бернулли с параметром $p$. Найденная вероятностная программа представлена на рис.~\ref{program:Bernoulli_samplers}.

\begin{figure}
\lstset{
  basicstyle=\normalsize,
  numbers=none,
  language=Lisp
}
\begin{lstlisting}
(lambda (par stack-id) (if (< (uniform-continuous 0.0 1.0) par) 1.0 0.0))

(lambda (par stack-id)
  (if (< 1.0 (safe-sqrt (safe-div par (safe-uc par (dec par))))) 1.0 0.0))
(lambda (par stack-id)
  (if (< 1.0 (safe-uc (safe-sqrt par) (+ par (cos par)))) 1.0 0.0))
\end{lstlisting}
\vspace{-4mm}
\caption{{\it (сверху)} Написанный человеком исходный код генератора случайных чисел, распределенных по закону Бернулли $\mathrm{Bernoulli}(p)$. {\it (внизу, две программы)} Выведенные исходные коды. Первая из двух выведенных вероятностных программ определяет настоящее семейство распределений Бернулли $\mathrm{Bernoulli}(p)$, параметризованное $p$. Вторая программа генерирует распределение, приближенное к распределению Бернулли, параметризованное $p$.}
\label{program:Bernoulli_samplers}
\end{figure}

На рис.~\ref{fig:found_approximators_as_probabilistic_programs} представлен выведенный исходный текст вероятностной программы для генерации элементов из Гамма-распределения $\mathrm{Gamma}(a, 1)$, параметризованного параметром $a$.

\begin{figure}
\lstset{
  basicstyle=\footnotesize,
  numbers=none,
  language=Lisp
}
\begin{tabular}{ m{0.60\textwidth} m{0.40\textwidth} }
\begin{lstlisting}{}{program:abc}
(lambda (par stack-id) (* (begin (define sym0 0.0)
    (exp (safe-uc -1.0 (safe-sqrt (safe-uc
    (safe-div (safe-uc 0.0 (safe-uc 0.0 3.14159))
      par) (+ 1.0 (safe-uc (begin (define sym2
      (lambda (var1 var2 stack-id) (dec var2)))
      (sym2 (safe-uc -2.0 (* (safe-uc 0.0 (begin
      (define sym4 (safe-uc sym0 (* (+ (begin
      (define sym5 (lambda (var1 var2 stack-id)
      (safe-div (+ (safe-log (dec 0.0)) -1.0) var1)))
      (sym5 (exp par) 1.0 0)) 1.0) 1.0)))
      (if (< (safe-uc par sym4) 1.0) sym0
      (safe-uc 0.0 -1.0)))) sym0))
      (safe-div sym0 (exp 1.0)) 0)) 0.0))))))) par))
\end{lstlisting} & \begin{lstlisting}{}{program:abc}
(lambda (stack-id)
  (* 2.0 (* (*
      (* -1.0 (safe-uc 0.0 2.0))
      (safe-uc (safe-uc 4.0
        (+ (safe-log 2.0) -1.0))
        (* (safe-div 2.0
          -55.61617747203855)
          (if (< (safe-uc
            (safe-uc
            27.396810474207317
             (safe-uc -1.0 2.0))
            2.0) 2.0)
            4.0 -1.0)))) -1.0)))
\end{lstlisting} \\
\end{tabular}
\vspace{-7mm}
\caption{Исходный код выведенных вероятностных программ для {\it (слева)} Гамма-распределения $\mathrm{Gamma}(a, 1)$ и {\it (справа)} для третьей эмпирической выборки индикаторов, используемых для рассмотрения заявок на выдачу кредита. Выборки, сгенерированные с помощью выполнения этих программ, расположены соответственно на рис.~\ref{fig:common_distributions_histograms} и на рис.~\ref{fig:real_world_data_approximation_histograms} {\it (последняя гистограмма из трех)}. Для экономии места исходный код был сокращен, где возможно; например, путем замены \texttt{(* 1.0 0.0)} на \texttt{0.0}.}
\label{fig:found_approximators_as_probabilistic_programs}
\end{figure}

\subsubsection{Обобщение произвольных эмпирических данных с помощью порождающего вероятностного программирования}

Мы также проверили наш метод на выводе порождающих моделей в виде вероятностных программ для объяснения настоящих, не синтетических, данных, аналитическое распределение которых неизвестно.

\begin{figure}
\vspace{-2mm}
\includegraphics[width=0.33\textwidth]{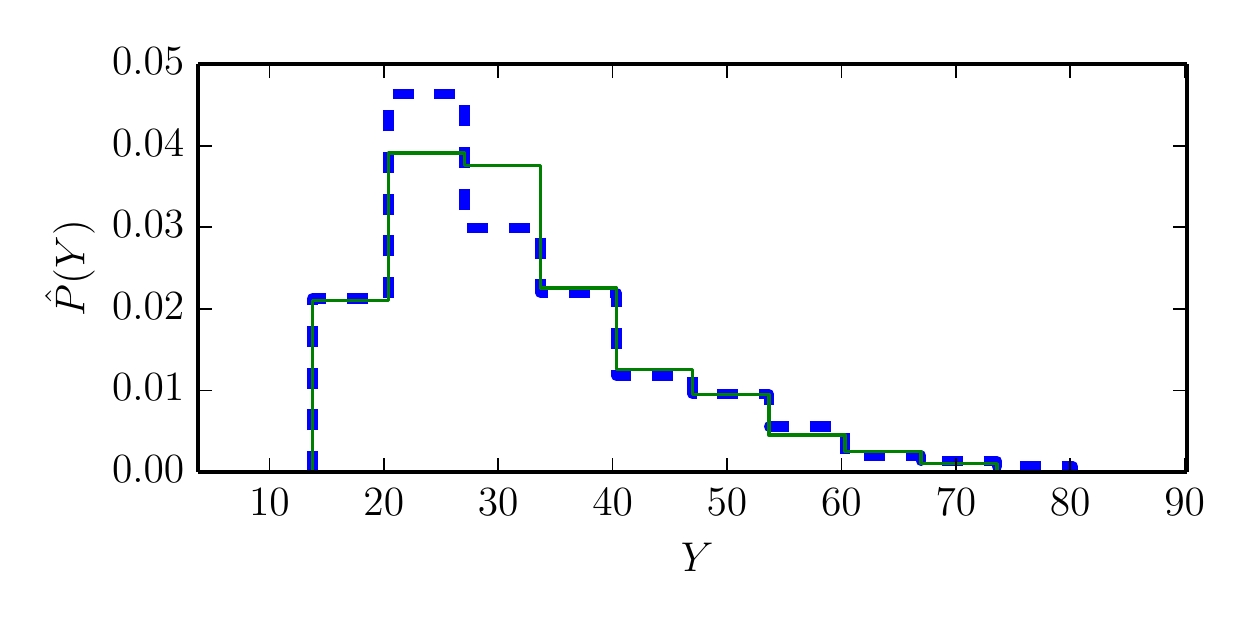}
\includegraphics[width=0.33\textwidth]{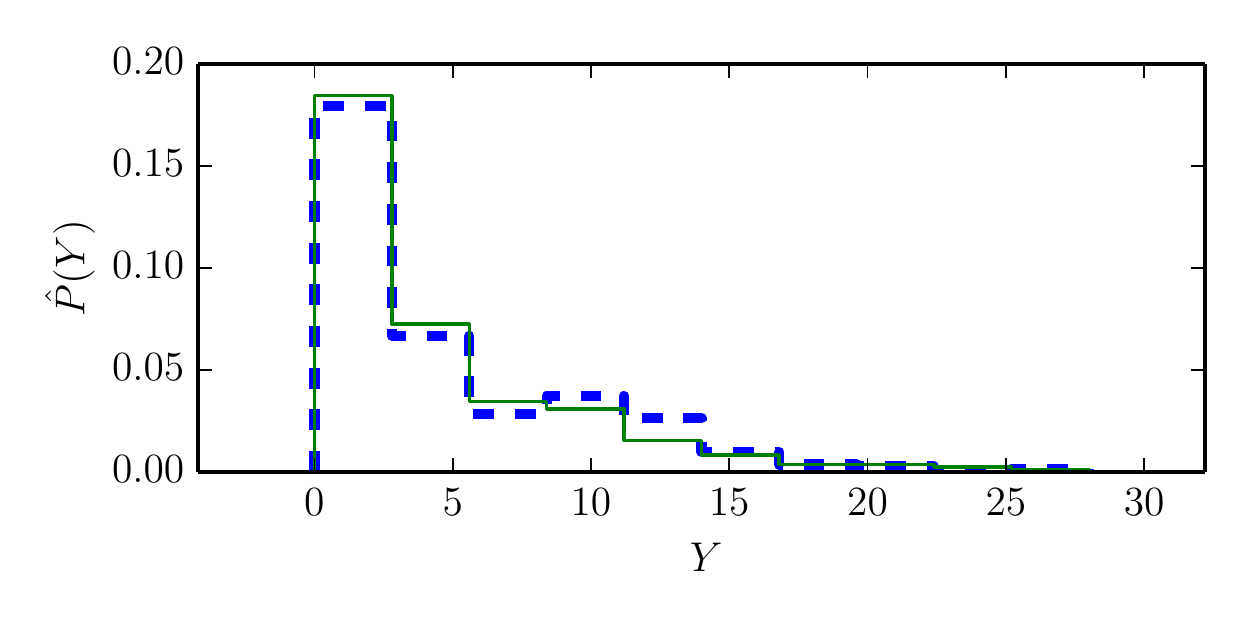}
\includegraphics[width=0.33\textwidth]{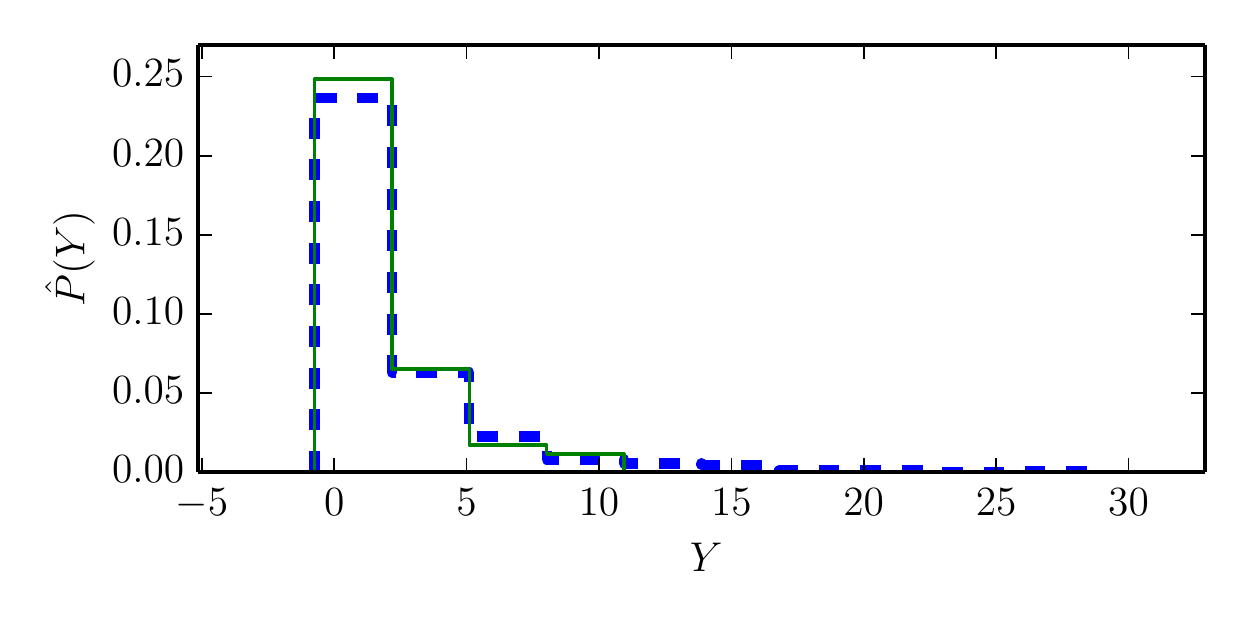}
\caption{Гистограммы {\it (зеленые сплошные линии)} выборок из распределений, определяемых найденными в процессе статистического вывода вероятностными программами для аппроксимации эмпирических данных {\it (синие пунктирные линии)} трех вещественных показателей из базы данных, используемой для анализа заявок на выдачу кредита.}
\label{fig:real_world_data_approximation_histograms}
\end{figure}

Для данного эксперимента мы выбрали три набора данных признаков обращающихся в банк клиентов для получения кредита \cite{quinlan1987simplifying,Bache+Lichman:2013}, и производили вывод вероятностной программы, используя для сравнения $p$-значения двухвыборочного теста Колмогорова-Смирнова аналогично тому, как мы использовали G-тест для дискретных распределений. Гистограммы выборок из лучших найденных вероятностных программ в сравнении с гистограммами истинных эмпирическими данными признаков клиентов приведены на рис.~\ref{fig:real_world_data_approximation_histograms}. Пример выведенной вероятностной программы показан на рис.~\ref{fig:found_approximators_as_probabilistic_programs}~{\it (справа)}.

\subsubsection{<<Компиляция>> вероятностных программ}

Генерация выборок из апостериорного распределения методом Монте-Карло по схеме Марковских цепей, особенно в случае Байесовского вывода, обычно достаточно дорогостояща. Под <<компиляцией>> вероятностных программ мы имеем в виду поиск вероятностных программ, априорное распределение которых согласовывалось бы точно или приблизительно с искомым апостериорным распределением.

В вероятностном программировании, где в общем случае генерация выборки из желаемого апостериорного распределения путем использования методом Монте-Карло или аналогичным им является единственным доступным средством за разумное время, эта проблема стоит еще острее.

В качестве самых предварительных результатов мы провели эксперимент, в рамках которого наша априорная модель была моделью бета-биномиального распределения (в несжатой форме) с априорным распределением на скрытый параметр $\theta \sim \mathrm{Beta}(1.0, 1.0)$, и использовали алгоритм Метрополиса-Гастингса для получения выборки $\hat{\mathcal{X}}$ из апостериорного распределения скрытого параметра $\theta$ с учетом проведения четырех успешных испытаний по схеме Бернулли $\mathrm{Bernoulli}(\theta)$. Соответствующая вероятностная программа представлена на рис.~\ref{program:CompilationCase}.

\begin{figure}
\lstset{
  basicstyle=\small,
  numbers=none,
  language=Lisp
}
\begin{tabular}{ m{0.37\textwidth} m{0.40\textwidth} }
\begin{lstlisting}{}{program:abc}
[ASSUME theta (beta 1.0 1.0)]
[OBSERVE (flip theta) True]
[OBSERVE (flip theta) True]
[OBSERVE (flip theta) True]
[OBSERVE (flip theta) True]
[PREDICT theta]
[PREDICT (flip theta)]
\end{lstlisting} & \begin{lstlisting}{}{program:abc}
[ASSUME theta (safe-beta 4.440 1.0)]
[ASSUME theta (safe-sqrt
                 (safe-beta (safe-log 11.602) 1.0))]
[ASSUME theta (safe-beta (safe-sqrt 27.810) 1.0)]

[ASSUME theta (beta 5.0 1.0)]
[PREDICT theta]
\end{lstlisting} \\
\end{tabular}
\vspace{-6mm}
\caption{{\it (слева)} Вероятностная модель бета-биномиального распределения (в несжатой форме) в виде вероятностной программы. В рамках эксперимента была поставлена задача найти формализацию апостериорного распределения скрытого параметра $\theta$ в виде другой вероятностной программы, чье априорное распределение будет совпадать или приближать данное апостериорное. {\it (справа, сверху)} Найденные вероятностные программы (т.е., результат <<компиляции>> вероятностных программ), априорное распределение которых приближает заданное апостериорное распределение. {\it (справа, внизу)} Записанный человеком исходный код вероятностной программы, чье априорное распределение совпадает с априорным распределением. В данном конкретном случае данная вероятностная программа может быть просто выведена аналитически, так как Бета-распределение является сопряженным к биномиальному, но в общем случае это нетривиальная задача.}
\label{program:CompilationCase}
\end{figure}

Затем мы использовали наш подход к выводу вероятностных программ, априорное распределение которых будет статистически схожим с полученной выборкой из желаемого апостериорного распределения. Примеры найденных вероятностных программ даны на рис.~\ref{program:CompilationCase}. В данном конкретном случае мы можем аналитически найти и записать в виде вероятностной программы апостериорное распределение, равное $\mathrm{Beta}(5.0, 1.0)$. Таким образом, полученные результаты показывают, что мы нашли хорошее приближение апостериорного распределения.

\subsection{Обобщение порождающего вероятностного программирования}

Как было отмечено ранее, порождающие вероятностные модели определяют совместное распределение $\left( T, X \right)$, часто задаваемое сначала с помощью распределения $\left( T \right)$, а затем с помощью условного распределения $\left(X\ |\ T\right)$. Порождающие вероятностные модели могут использоваться либо для генерации выборок напрямую из $\left(T, X\right)$, либо в качестве промежуточного этапа для нахождения условного распределения скрытых параметров $\left(T\ |\ X = \hat{x}\right)$, обычно с помощью приближенных методов статистического вывода или оптимизации.

В рамках наших экспериментов, описанных выше, мы решали задачу вывода вероятностной программы, которая определяет вероятностную модель (в общем случае параметризованную), априорное распределение которой соответствует или приближает искомое распределение, заданное аналитически, или в виде выборки, или в виде дорогостоящего генератора элементов выборки.

На языке вероятностного программирования Church, поддерживающего проведение статистического запроса внутри другого статистического запроса, данная задача может быть сформулирована следующим образом (подобно тому, как мы ее формулировали в Anglican/Venture, см.~\ref{fig:inferring_std_normal_via_moments} и~\ref{fig:inferring_bernoulli_via_gtest}):

\lstset{
  numbers=left,
  language=Lisp
}
\begin{lstlisting}{}{program:abc}
(query
  (define program-text (productions INPUT-TYPES OUTPUT-TYPE))
  (define program
    (eval (list `lambda `(arg1 ... arg N) program-text)))
  program-text
  (and
    (noisy-distance-equal
      (get-statistics
        (multiple-query
          ; No defines.
          (program arg1_1 ... arg1_N)
          true))
      expected-statistics1)
    ...
    (noisy-distance-equal
      (get-statistics
        (multiple-query
          ; No defines.
          (program argM_1 ... argM_N)
          true))
      expected-statisticsM)))
\end{lstlisting}
где \texttt{(query defines expression predicate)} --- задание запроса в Church (см.~\cite{goodman2008church}), \texttt{INPUT-TYPES} --- перечисление типов входных аргументов, \texttt{OUTPUT-TYPE} --- тип выходного значения процедуры, \texttt{noisy-distance-equal} --- функция сравнения статистик, \texttt{get-statistics} --- функция извлечения статистик из распределения, \texttt{multiple-query} --- аналог \texttt{query}, возвращающий несколько элементов из распределения вместо одного элемента, \texttt{expected-statistics...} --- M значений статистик искомого распределения.

Автор считает, что более общо, для нахождения вероятностной модели, служащей вспомогательным инструментом в нахождении условного апостериорного распределения $P(T\ |\ X = \hat{x})$, мы можем быть заинтересованы в постановке задачи следующим образом:

\lstset{
  numbers=left,
  language=Lisp
}
\begin{lstlisting}{}{program:abc}
(query
  (define latent-variables-creator (latent-productions LATENT-TYPES))
  (define program-text
    (productions LATENT-TYPES INPUT-TYPES OUTPUT-TYPE))
  (define program
    (eval (list `lambda `(arg1 ... argN) program-text)))
  (list latent-variables-creator program-text)
  (and
    (noisy-latents-equal
      (multiple-query
        (define latent-variables (latent-variables-creator))
        latent-variables
        (noisy-outputs-equal
          (program latent-variables arg1_1 ... arg1_N) output1))
      expected-latents1)
    ...
    (noisy-latents-equal
      (multiple-query
        (define latent-variables (latent-variables-creator))
        latent-variables
        (noisy-outputs-equal
          (program latent-variables argM_1 ... argM_N) outputM))
      expected-latentsM)))
\end{lstlisting}
где \texttt{latent-productions} --- правила порождения объекта, представляющего скрытые параметры типов \texttt{LATENT-TYPES}, <<лениво>> создаваемые с помощью процедуры \texttt{mem} (см. \cite{goodman2008church}); \texttt{latent-variables} --- объект, представляющий скрытые параметры; \texttt{arg...\_...} --- M значений аргументов, \texttt{output...} --- M значений наблюдаемых данных; \texttt{expected-latents...} --- M значений ожидаемых скрытых параметров.

Таким образом, в рамках данной формулировки мы ищем вероятностную программу, содержащую: скрытые переменные, определяемые \texttt{latent-variables}; наблюдаемые переменные, определяемые выполнением \texttt{program}; и вывод производится с учетом M тренировочных троек $$\{a_i, \hat{t}_i, \hat{x}_i\},$$ где $a_i$ --- дополнительная параметризация, $\hat{x}_i$ --- значения наблюдаемых переменных, $\hat{t}_i$ --- ожидаемые значения скрытых переменных.

Следует отметить, что так как в настоящее время обобщенный статистический вывод с использованием вероятностных языков программирования производится достаточно медленно, то решение практических задач таким образом пока не имеет смысла. С другой стороны, с дальнейшим развитием \cite{wood2014new,venture} обобщенных методов статистического вывода и других решений (например, оптимизации или поиска), с использовании более продуманных алгоритмов программной реализации и структур данных \cite{venture,Paige-ICML-2014}, со специализированным аппаратным обеспечением \cite{tenenbaum2008stochastic,jonas2014stochastic}, автоматизированная генерация и использование порождающих вероятностных моделей может успешно и продуктивно применяться на практике.

\clearpage

\begingroup
\renewcommand{\section}[2]{}
\section*{ЗАКЛЮЧЕНИЕ}
\addcontentsline{toc}{section}{Заключение}
\begin{center}
{\centering \Large \bf ЗАКЛЮЧЕНИЕ}
\end{center}
\vspace{5mm}
\endgroup

В первой части данной работы очень кратко и реферативно представлено введение в вероятностное программирование на примере языков Church/Venture/Anglican, первое подобное введение на русском языке, насколько известно автору.

Во второй части данной работы представлен подход к порождению вероятностных программ, обобщающих распределения, представленные либо в виде выборки, либо в виде аналитического представления (например, в виде значения статистик), причем мета-вероятностная модель вероятностных моделей также записывается в виде вероятностной программы. Используя данный подход, были получены предварительные положительные результаты статистического вывода из апостериорного распределения искомых вероятностных программ методами Монте-Карло по схеме цепей Маркова, в том числе была автоматически выведена вероятностная программа, генерирующая элементы выборки всего семейства распределений Бернулли.

\clearpage

\begingroup
\renewcommand{\section}[2]{}
\section*{СПИСОК ИСПОЛЬЗОВАННЫХ ИСТОЧНИКОВ}
\addcontentsline{toc}{section}{Список использованных источников}
\begin{center}
{\centering \Large \bf СПИСОК ИСПОЛЬЗОВАННЫХ ИСТОЧНИКОВ}
\end{center}
\vspace{5mm}
\endgroup

\begingroup
\renewcommand{\section}[2]{}%
\renewcommand{\addcontentsline}[3]{}
\bibliography{references}
\endgroup

\end{document}